\documentclass[review]{elsarticle}
\usepackage[mathscr]{euscript}
\usepackage{array}
\usepackage{eqnarray,amsmath}
\usepackage{hyperref}
\usepackage{graphicx}
\usepackage{colortbl}
\usepackage{tabularx}
\usepackage{multirow}
\usepackage{booktabs}
\usepackage{cases}
\usepackage{Algorithm}
\usepackage[noend]{Algorithmic}

\journal{Journal of \LaTeX\ Templates}

\begin{document}

\begin{frontmatter}

\title{Deep Learning for Saliency Prediction in Natural Video}

\author{Souad CHAABOUNI\fnref{mymainaddress,mysecondaryaddress}}
\ead{souad.chaabouni@labri.fr}

\author{Jenny BENOIS-PINEAU\fnref{mymainaddress}}
\ead{benois-p@labri.fr}

\author{Ofer HADAR\fnref{mythirdaddress}}
\ead{hadar@bgu.ac.il}

\author{Chokri BEN AMAR\fnref{mysecondaryaddress}}
\ead{chokri.benamar@ieee.org}

\address[mymainaddress]{Universit\'{e} de Bordeaux, Laboratoire Bordelais de Recherche en Informatique, B\^{a}timent $\ \ $ A30, F-33405 Talence cedex, France}
\address[mysecondaryaddress]{Sfax university, Research Groups in Intelligent Machines, National Engineering School of $\ $Sfax (ENIS), Tunisia }
\address[mythirdaddress]{Communication Systems Engineering department, Ben Gurion University of the Nagev $\ \ \ \ $}

\begin{abstract}
The purpose of this paper is the detection of salient areas in natural video by using the new deep learning techniques. 
Salient patches in video frames are predicted first.
Then the predicted visual fixation maps are built upon them.  We design the deep architecture on the basis of CaffeNet implemented 
with Caffe toolkit. We show that changing
the way of data selection for optimisation of network parameters, we can save computation cost up to $12$ times. 
We extend deep learning approaches for saliency prediction in still images with RGB values to specificity of video using the sensitivity 
of the human visual system to residual motion. 
Furthermore, we complete primary colour pixel values by contrast features proposed in classical visual attention prediction models.  
The experiments are conducted on two publicly available datasets. The first is IRCCYN video database containing $31$ videos with an 
overall amount of $7300$ frames and eye fixations 
of $37$ subjects.
The second one is HOLLYWOOD2 provided $2517$ movie clips with the eye fixations of $19$ subjects.
On IRCYYN dataset, the accuracy obtained is of $89.51\% $.
On HOLLYWOOD2 dataset, results in prediction of saliency of patches show the  improvement up to $2\%$ with regard to RGB use only. 
The resulting accuracy of $76,6\%$ is obtained. 
The AUC metric in comparison of predicted saliency maps with visual fixation maps shows the increase up to $16\%$ on a sample of video 
clips from this dataset.  

\end{abstract}


\end{frontmatter}

\section{Introduction}

Deep learning has emerged as a new field of research in machine learning, providing learning at multiple levels of abstraction for mining the data 
such as images, sound and 
text\cite{New51}. Although, it is hierarchically created usually on the basis of neural networks, deep learning presents a philosophy to model the
complex relationships between data \cite{New2}, 
\cite{New18}. Since recently, deep learning has become the most exciting field which attracts many researchers. First, to understand the new deep networks 
in itself (\cite{New13},\cite{New33}, \cite{New34}, \cite{New35}, \cite{New37}, \cite{New38}, \cite{New39},),
such as the important question in building a deep convolutional network, is the optimization of pooling layer \cite{New14}. 
Second, to use that deep network in their original domain such as object recognition \cite{New40}, \cite{New9}, \cite{New36}, \cite{New42}, multi-task learning \cite{New40}. 
As a definition,
neural networks are generally multilayer generative networks formed to maximize the probability 
of input data with regard to target classes.

The predictive power of Deep Convolutional Neural Networks (CNN) is interesting for the use in the problem of prediction of visual attention in visual content, i.e. saliency of the latter.
Indeed, several saliency models have been proposed in various fields of research such as psychology and neurobiology, which are based on the feature integration theory
(\cite{New19}, \cite{New23}, \cite{New24}, \cite{New25}, \cite{New26}, \cite{New27}, \cite{New28}, \cite{New29}, \cite{New30}, \cite{New31}, \cite{New32}, \cite{New8},..). 
These research models the so-called "bottom-up" saliency with the theory that suggests the visual characteristics of low-level as luminance, color, orientation and movement 
to provoke human gaze attraction \cite{New48}. The "bottom-up" models have been extensively studied in the literature \cite{New48}. They suffer from insufficiency of 
low-level features in the feature integration theory framework, especially when 
the scene contains significant content and semantic objects. In this case, the so-called "top-down" attention \cite{New49} becomes prevalent, 
the human subject observes visual content  progressively  with increasing the time of looking of the visual sequence.
Supervised machine learning techniques help in detection of salient regions in images predicting  attractors on the basis of seen data\cite{New6}.
Various recent research is directed towards the creation of a basic deep learning
model ensuring the detection of salient areas. We can cite here \cite{New4}, \cite{New5} and \cite{New6}. While a significant effort has
been already done for building such models from still images, very few models have been built for video content for saliency prediction with supervised learning \cite{New52}.
It has a supplementary dimension: the temporality expressed by apparent motion in the image plane.

In this paper, we present a new approach with Deep CNN that ensures the learning of salient areas in order to predict the saliency maps in videos.
The paper is organized as follows. Section 2 describes the related work of  different deep learning models used to detect  salient areas
in images or to classify images by content.
Section 3 presents  our proposed method for detection of salient regions with a deep learning approach. Pixel-wise computation of predicted visual attention/saliency maps is then introduced.
In section 4 we present results and comparison with reference methods of the state-of-the-art. Section 5 concludes the paper and outlines the perspectives of this research.

\section{Related work}
Deep learning architectures which have been recently proposed for the prediction of salient areas in images differ essentially by the quantity of convolution and pooling layers, by the input data,  by pooling strategies, by the nature of the final classifiers and the loss functions to optimize, but also by the 
formulation of the problem. The attempt to predict visual attention reveals the binary classification problem of areas in images as "salient" and "non-salient". It corresponds to the visual experiment with free instructions, when the subjects are simply asked to look at the content. Shen \cite{New4} proposed a deep learning model to extract salient areas in images. It allows firstly to learn the relevant characteristics of the saliency of natural images, and secondly to predict the eye fixations on 
objects with semantic content. The proposed model is formed by three layer sequences of "filtering" and "pooling", followed by a layer of linear SVM classifier providing ranked "salient"
or "non-salient" 
regions of the input image. With the filtering by sparse coding and the max pooling, this model approximates human gaze fixations.

In Simonyan's work \cite{New5} the saliency of image pixels is defined with regard to a given class in image taxonomy as a relevance of the image for the class.
Therefore the classification problem is multi-class, and can be expressed as a "task-dependent" visual experiment, where the subjects are asked to look for an object of a given 
taxonomy in the images. The creation of the saliency map for each class using deep CNN with optimisation of parameters by stochastic gradient descent, presents the challenge of this
research \cite{New5}. After a step of generating the map that maximizes the score of the specific class, the saliency map of each class is defined by the amplitude of the weight calculated from the convolution
network with a single layer.

The learning model of salient areas  proposed by Vig \cite{New6} tackles prediction of saliency of pixels for a human visual system (HVS) and corresponds to a free-viewing visual experience. It comprises two phases. First, a random bank of uniform filters is used to generate multiple representations of localized input images.
The second phase provides the combination of different localized representations. The training step is summarized by the random token, from the combined representation of each image, 
of regions composed of ten pixels, and granted to each region a saliency class by reference to the density fixations map.
The integration of this set in a SVM classifier allows the creation of the learning model. 
The learning model of salient areas is composed by the SVM trained on the combination of feature maps that are obtained 
using of different architectures of deep network.

In our work we also seek for predicting saliency of image regions for HVS.
While in  \cite{New5} only primary RGB pixel values are taken for class-based saliency prediction, we use several combinations of primary (input) 
features such as residual motion and primary spatial features, 
inspired by feature integration theory as in \cite{New15}, \cite{New8}, \cite{New16}, \cite{New32}. The sensitivity of HVS 
to residual motion in dynamic visual scenes is used for saliency prediction in video \cite{New32}. For  training of deep CNN in our two class classification problem  we use human fixations maps as in \cite{New6} to select positive and negative samples. 

\section{Prediction of visual saliency with deep CNN}

Hence we design a deep CNN to classify regions in video frames into two classes salient and non-salient. Then on the basis of these 
classifications, a visual fixation map will be predicted. 
Before describing the architecture of our proposed deep CNN, we introduce the definition of saliency of regions and explain 
how we extract positive and negative  examples for training the CNN. 

\subsection{Extraction of salient and non-salient patches.}
We define a \textit{salient} patch in a video frame on the basis of interest expressed by the subjects. 
The latter is measured by the magnitude of a visual attention map built upon gaze fixations which are recorded during a psycho-visual 
experiment in free-viewing conditions. The maps are built by the method of Wooding \cite{New12}. Such a map represents a multi-Gaussian 
surface normalized by its global maximum. To train the network it is necessary to extract
salient and non-salient patches from training video frames with available Wooding maps.  
A squared patch ${P}$ of parametrized size ${t \times t}$ is considered "salient" if the visual attention map ${W}$ value in its center is above a threshold. A patch $P$ is a vector in $R^{t\times t \times n}$, where $n$ stands for the quantity of primary feature maps serving as an input to the CNN. In case when RGB planes of a colour video sequences are used, $n=3$. The choice of the parameter $t$ obviously depends on the resolution of video, 
but also is constrained by the computational capacity to process a huge amount of data. In this work we considered $t=100$ for SD video.  
More formally, a binary label is associated with pixels ${X}$ of each patch ${P_i}$ using equation \eqref{eq:EqSaliencyPatch}:

\begin{equation}
\label{eq:EqSaliencyPatch}
 l(X)= \left\{
\begin{array}{l}
  1 \ \ \ \ if  \ \ \ \  W(x_{0,i}, y_{0,i}) \geq \tau_J \\
  0 \ \ \ \ otherwise 
\end{array}
\right.
\end{equation}
with $(x_{0,i}, y_{0,i})$ the coordinates of the center of the patch. 
We select a set of thresholds, starting by the global maximum value of the normalized attention map and then relax threshold as in equation\eqref{eq:Threshold}:
\begin{equation}
\label{eq:Threshold}
\left\{
\begin{array}{l}
  \tau_0 = \max (W(x,y),0) \\
  \tau_{(j+1)}=\tau_j - \epsilon \tau_j
\end{array}
\right. 
\end{equation}
Here $0<\epsilon<1$ is a relaxation parameter, $j=0,\cdots,J$,  and $J$ limits the relaxation of  saliency. It was chosen experimentally as $J=5$, while $\epsilon=0.04$. 
 
\begin{center}

\begin{figure}
\begin{center}
\includegraphics[width=0.9\linewidth]{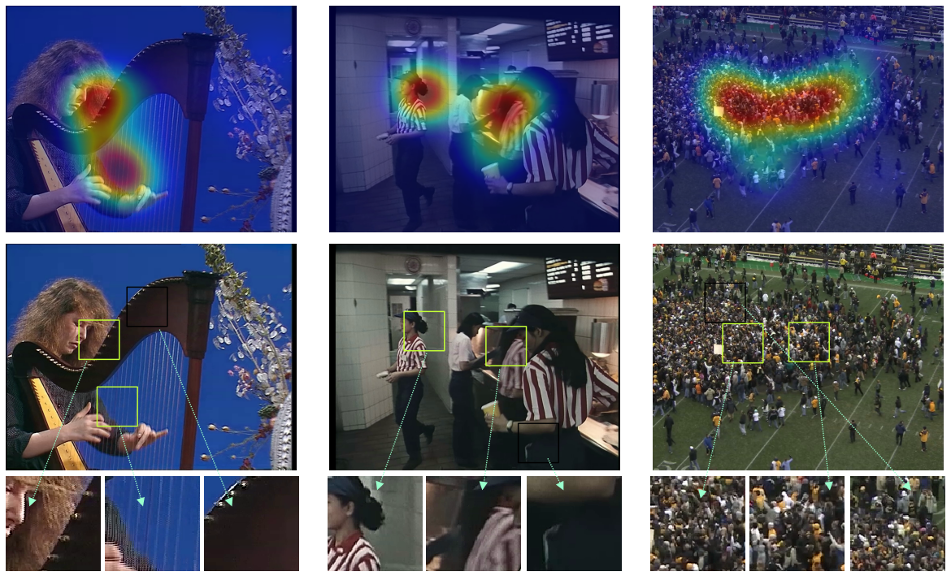}
\end{center}
\caption{\label{ExtractPatches}Extraction of salient patches for training}
\end{figure}
\end{center}

In such a manner, salient patches are progressively selected up to non-salient areas, where non-salient patches are extracted randomly. The process of extraction of salient patches
in the frames of training videos is illustrated in figure \ref{ExtractPatches}.    

The tables \ref{tab1}, \ref{tab2} present the group of salient patches on the left and non-salient patches on the right, each row presents some examples of patches taken from each frame 
of video sequence denoted by "SRC" in  IRCCYN \footnote{available in  ftp://ftp.ivc.polytech.univ-nantes.fr/IRCCyN\_IVC\_Eyetracker\_SD\_2009\_12/} dataset, and "actioncliptrain" in 
the HOLLYWOOD\footnote{available in http://www.di.ens.fr/$\sim$laptev/actions/hollywood2/} data set.

\begin{table}
\caption{ Training data from IRCCYN data base} 
\label{tab1}
\begin{tabular}{ ccccc||ccccc }
\includegraphics[width=0.09\linewidth]{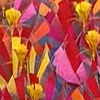} & \includegraphics[width=0.09\linewidth]{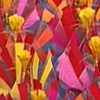} & \includegraphics[width=0.09\linewidth]{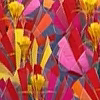} & \includegraphics[width=0.09\linewidth]{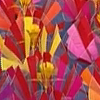} & \includegraphics[width=0.09\linewidth]{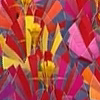} & \includegraphics[width=0.09\linewidth]{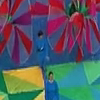} \includegraphics[width=0.09\linewidth]{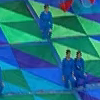} & \includegraphics[width=0.09\linewidth]{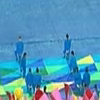} & \includegraphics[width=0.09\linewidth]{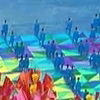} & \includegraphics[width=0.09\linewidth]{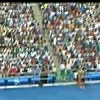}  \\ 
\includegraphics[width=0.09\linewidth]{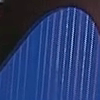} & \includegraphics[width=0.09\linewidth]{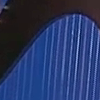} & \includegraphics[width=0.09\linewidth]{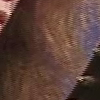} & \includegraphics[width=0.09\linewidth]{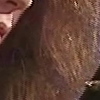} & \includegraphics[width=0.09\linewidth]{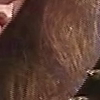} & \includegraphics[width=0.09\linewidth]{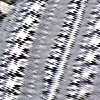} \includegraphics[width=0.09\linewidth]{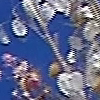} & \includegraphics[width=0.09\linewidth]{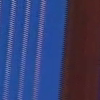} & \includegraphics[width=0.09\linewidth]{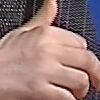} & \includegraphics[width=0.09\linewidth]{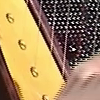}  \\ 
\includegraphics[width=0.09\linewidth]{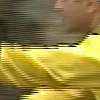} & \includegraphics[width=0.09\linewidth]{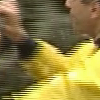} & \includegraphics[width=0.09\linewidth]{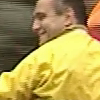} & \includegraphics[width=0.09\linewidth]{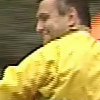} & \includegraphics[width=0.09\linewidth]{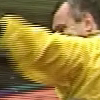} & \includegraphics[width=0.09\linewidth]{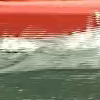} \includegraphics[width=0.09\linewidth]{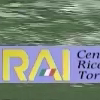} & \includegraphics[width=0.09\linewidth]{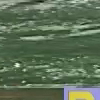} & \includegraphics[width=0.09\linewidth]{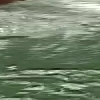} & \includegraphics[width=0.09\linewidth]{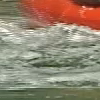}  \\ 
\includegraphics[width=0.09\linewidth]{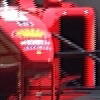} & \includegraphics[width=0.09\linewidth]{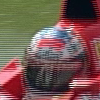} & \includegraphics[width=0.09\linewidth]{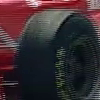} & \includegraphics[width=0.09\linewidth]{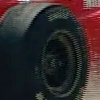} & \includegraphics[width=0.09\linewidth]{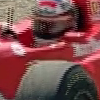} & \includegraphics[width=0.09\linewidth]{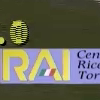} \includegraphics[width=0.09\linewidth]{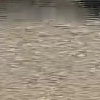} & \includegraphics[width=0.09\linewidth]{images/trainS/0111_0.png} & \includegraphics[width=0.09\linewidth]{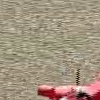} & \includegraphics[width=0.09\linewidth]{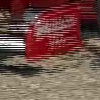}  \\ 
\includegraphics[width=0.09\linewidth]{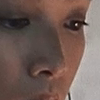} & \includegraphics[width=0.09\linewidth]{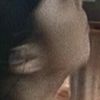} & \includegraphics[width=0.09\linewidth]{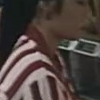} & \includegraphics[width=0.09\linewidth]{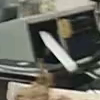} & \includegraphics[width=0.09\linewidth]{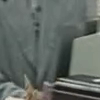} & \includegraphics[width=0.09\linewidth]{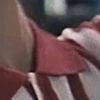} \includegraphics[width=0.09\linewidth]{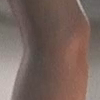} & \includegraphics[width=0.09\linewidth]{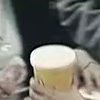} & \includegraphics[width=0.09\linewidth]{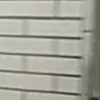} & \includegraphics[width=0.09\linewidth]{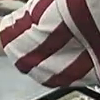}  \\ 
\includegraphics[width=0.09\linewidth]{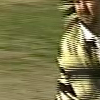} & \includegraphics[width=0.09\linewidth]{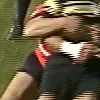} & \includegraphics[width=0.09\linewidth]{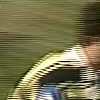} & \includegraphics[width=0.09\linewidth]{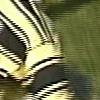} & \includegraphics[width=0.09\linewidth]{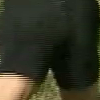} & \includegraphics[width=0.09\linewidth]{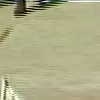} \includegraphics[width=0.09\linewidth]{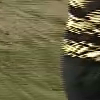} & \includegraphics[width=0.09\linewidth]{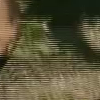} & \includegraphics[width=0.09\linewidth]{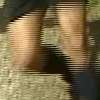} & \includegraphics[width=0.09\linewidth]{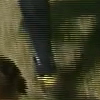}  \\ 
\multicolumn{5}{c||}{ \tiny{Salient patch $\#SRC_{\{1..6\}}$}}  & \multicolumn{5}{c}{\ \ \ \ \ \ \tiny{Non-salient patch $\#SRC_{\{1..6\}}$ }} \\
\end{tabular}   
\end{table}

\begin{table}
\caption{ Training data from HOLLYWOOD data base} 
\label{tab2}
\begin{tabular}{ ccccc||ccccc }
\includegraphics[width=0.09\linewidth]{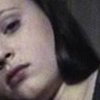} & \includegraphics[width=0.09\linewidth]{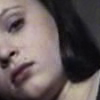} & \includegraphics[width=0.09\linewidth]{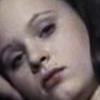} & \includegraphics[width=0.09\linewidth]{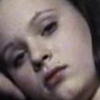} & \includegraphics[width=0.09\linewidth]{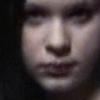} & \includegraphics[width=0.09\linewidth]{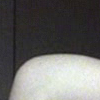} \includegraphics[width=0.09\linewidth]{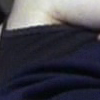} & \includegraphics[width=0.09\linewidth]{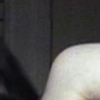} & \includegraphics[width=0.09\linewidth]{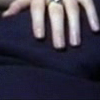} & \includegraphics[width=0.09\linewidth]{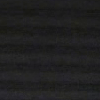}  \\ 
\includegraphics[width=0.09\linewidth]{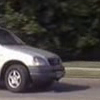} & \includegraphics[width=0.09\linewidth]{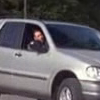} & \includegraphics[width=0.09\linewidth]{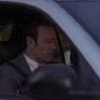} & \includegraphics[width=0.09\linewidth]{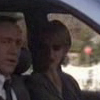} & \includegraphics[width=0.09\linewidth]{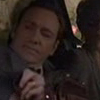} & \includegraphics[width=0.09\linewidth]{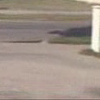} \includegraphics[width=0.09\linewidth]{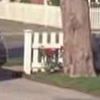} & \includegraphics[width=0.09\linewidth]{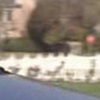} & \includegraphics[width=0.09\linewidth]{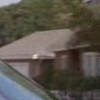} & \includegraphics[width=0.09\linewidth]{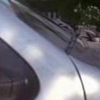}  \\ 
\includegraphics[width=0.09\linewidth]{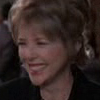} & \includegraphics[width=0.09\linewidth]{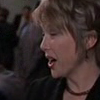} & \includegraphics[width=0.09\linewidth]{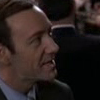} & \includegraphics[width=0.09\linewidth]{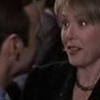} & \includegraphics[width=0.09\linewidth]{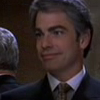} & \includegraphics[width=0.09\linewidth]{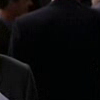} \includegraphics[width=0.09\linewidth]{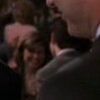} & \includegraphics[width=0.09\linewidth]{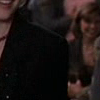} & \includegraphics[width=0.09\linewidth]{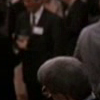} & \includegraphics[width=0.09\linewidth]{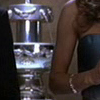}  \\ 
\includegraphics[width=0.09\linewidth]{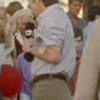} & \includegraphics[width=0.09\linewidth]{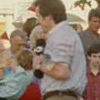} & \includegraphics[width=0.09\linewidth]{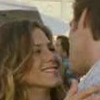} & \includegraphics[width=0.09\linewidth]{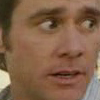} & \includegraphics[width=0.09\linewidth]{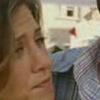} & \includegraphics[width=0.09\linewidth]{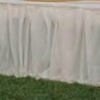} \includegraphics[width=0.09\linewidth]{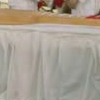} & \includegraphics[width=0.09\linewidth]{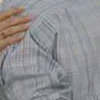} & \includegraphics[width=0.09\linewidth]{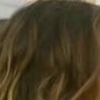} & \includegraphics[width=0.09\linewidth]{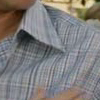}  \\ 
\includegraphics[width=0.09\linewidth]{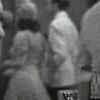} & \includegraphics[width=0.09\linewidth]{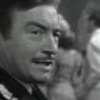} & \includegraphics[width=0.09\linewidth]{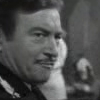} & \includegraphics[width=0.09\linewidth]{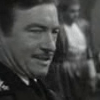} & \includegraphics[width=0.09\linewidth]{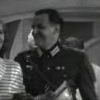} & \includegraphics[width=0.09\linewidth]{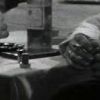} \includegraphics[width=0.09\linewidth]{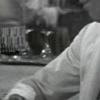} & \includegraphics[width=0.09\linewidth]{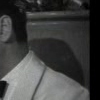} & \includegraphics[width=0.09\linewidth]{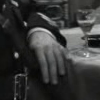} & \includegraphics[width=0.09\linewidth]{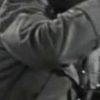}  \\ 
\includegraphics[width=0.09\linewidth]{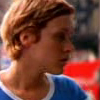} & \includegraphics[width=0.09\linewidth]{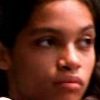} & \includegraphics[width=0.09\linewidth]{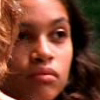} & \includegraphics[width=0.09\linewidth]{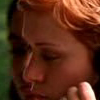} & \includegraphics[width=0.09\linewidth]{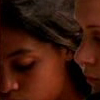} & \includegraphics[width=0.09\linewidth]{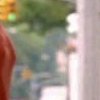} \includegraphics[width=0.09\linewidth]{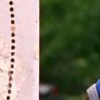} & \includegraphics[width=0.09\linewidth]{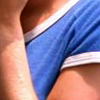} & \includegraphics[width=0.09\linewidth]{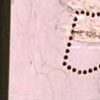} & \includegraphics[width=0.09\linewidth]{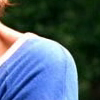}  \\ 
\multicolumn{5}{c||}{ \tiny{Salient patch $\#actioncliptrain_{\{1,2,11,105,171,596\}}$}}  & \multicolumn{5}{c}{\ \ \ \ \ \ \tiny{Non-salient patch $\#actioncliptrain_{\{1,2,11,105,171,596\}}$ }} \\
\end{tabular}   
\end{table}

\subsection{Primary feature maps for saliency prediction in video}
On the contrary to still natural images where  saliency is "spatial",  based on color contrasts, saturation contrasts, intensity contrasts $\cdots$, the saliency of the video is  also based on the motion information of the objects with regard to the background. 
Therefore, in the following we present primary motion features we consider and then briefly describe spatial primary features (colours, contrasts) we use.

\subsubsection{Motion feature maps}
Visual attention is not attracted by the motion in general, but by the difference between the global motion in the scene, expressing the camera work, and the "local" motion, that one of a moving object. This difference is called the 
"residual motion"\cite{New28}. To create the feature map of residual motion in  videos, we used the model developed in \cite{New10}, \cite{New28}, \cite{New46}. 
This model allows the calculation of the residual motion in three steps:
the optical flow estimation $\overrightarrow{M_c}(x,y)$,
the estimation of the global motion $\overrightarrow{M_\theta}(x,y)$, from optical flow accordingly to the first order complete affine model $\theta$ and finally, the computation of residual motion according to equation\eqref{eq:Residualmotion}:
\begin{equation}
\label{eq:Residualmotion}
\overrightarrow{M_r}(x,y) = \overrightarrow{M_\theta}(x,y) - \overrightarrow{M_c}(x,y) 
\end{equation}
The sensitivity of HVS to motion is selective. Daly \cite{New17} proposes a non-linear model of sensitivity accordingly to the speed of motion. In our work, we use a simplified version: the primary motion feature is the magnitude of residual motion \eqref{eq:Residualmotion} in a 
given pixel, and leave the decision on the saliency of the patch to the CNN classifier.
For spatial primary features we  resort  to the work in \cite{New10} which yeilds coherent results accordingly to our studies in \cite{New32}.

\subsubsection{Primary spatial features}
\label{PrimarySpatialFeatures}
The choice of features from \cite{New10} is conditioned by relatively low computational cost and their good performance we have stated in \cite{New32}. 
The authors propose seven color contrast descriptors. As the color space 'Hue Saturation Intensity' (HSI) is more
appropriate to describe the perception and color interpretation by humans, the descriptors of the
spatial saliency are built in this color space. Five of these seven local descriptors depend on
the value of the hue, saturation and/or intensity of the pixel. These values are determined for
each frame $I$ of a video sequence, from a saturation factor $f^{sat}$ and an intensity factor $f^{int}$,
calculated using the equations \eqref{eq:saturation1},\eqref{eq:saturation2}:
\begin{equation}
\label{eq:saturation1}
f^{sat}(I,i,j)=\frac{Sat(I,i)+Sat(I,j)}{2} \times (k_{min} + (1 - k_{min}) \cdot Sat(I,i))
\end{equation}
\begin{equation}
\label{eq:saturation2}
f^{int}(I,i,j)=\frac{Int(I,i)+Int(I,j)}{2} \times (k_{min} + (1 - k_{min}) \cdot Int(I,i))
\end{equation}
Here $Sat(I,i)$ is the saturation of the pixel $i$ at coordinates $(x_i, y_i)$ and the value at $Sat(I,j)$ is
the saturation of the pixel at coordinates $(x_j, y_j)$ adjacent to the pixel $i$. The constant $k_{min}=0,21$
sets the minimum value for the protection of the interaction of pixel $i$ when the saturation
approaches zero \cite{New10}. Contrast descriptors are calculated by equations (\ref{eq:ColorContrast}$\dots$ \ref{eq:IntensityContrast}):

1. \textit{color} \textit{contrast}: the first input of the saliency of a pixel is obtained from the two factors of
saturation and intensity. This descriptor $X_1 (I, i)$ is calculated for each pixel $i$ and its eight
connected neighbors $j$ of the frame $I$, as in equation\eqref{eq:ColorContrast}:
\begin{equation}
\label{eq:ColorContrast}
X_1 (I,i)= \sum_{j \in \eta_i}  f^{sat}(I,i,j) \cdot f^{int}(I,i,j)
\end{equation}

2. \textit{hue} \textit{contrast}: a hue angle difference on the color wheel can produce a contrast. In other words,
this descriptor is related to the pixels having a hue value far from their neighbors (the largest
angle difference value is equal to $180^{\circ}$), see equation \eqref{eq:HueContrast}:
\begin{equation}
\label{eq:HueContrast}
X_2 (I,i)= \sum_{j \in \eta_i}  f^{sat}(I,i,j) \cdot f^{int}(I,i,j) \cdot \Delta^{hue}(I,i,j)
\end{equation}
The difference in color $\Delta^{hue}$ between the pixel $i$ and its neighbor $j=8$ is calculated accordingly to equations \eqref{eq:Delta1} and \eqref{eq:Delta2} :
\begin{eqnarray}
\label{eq:Delta1}
 \Delta^{hue}= 
 \begin{cases}
 \Delta^{\mu}(I,i,j)\ \ \ \ \ \ \ \ \ \ \ \ \     if\ \Delta^{\mu}(I,i,j) \leq 0.5 \\
 1-\Delta^{\mu}(I,i,j)\ \ \ \ \ \ else
 \end{cases}
\end{eqnarray}
\begin{equation}
\label{eq:Delta2}
 \Delta^{\mu}(I,i,j)=\vert Hue(I,i) - Hue(I,j) \vert
\end{equation}

3. \textit{contrast} \textit{of} \textit{opponents}: the colors located on the opposite sides of the hue wheel creating a very
high contrast. An important difference in tone level will make the contrast between active color ($hue<0,5 \simeq 180^\circ$)
and passive, more salient. This contribution to the salience of the pixel $i$ is defined by equation \eqref{eq:OpponetContrast}:
\begin{eqnarray}
\label{eq:OpponetContrast}
\begin{cases}
X_3 (I,i)= \sum_{j \in \eta_i}  f^{sat}(I,i,j) \cdot f^{int}(I,i,j) \cdot \Delta^{hue}(I,i,j) \\
if \ Hue(I,i)<0.5\ and\ Hue(I,j)\geq 0.5
\end{cases}
\end{eqnarray}

4. \textit{contrast} \textit{of} \textit{saturation}: occurs when low and high color saturation regions are close. Highly
saturated colors tend to attract visual attention, unless a low saturation region is surrounded by a
very saturated area. It is defined by equation \eqref{eq:SaturationContrast}:
\begin{equation}
\label{eq:SaturationContrast}
X_4 (I,i)= \sum_{j \in \eta_i}  f^{sat}(I,i,j) \cdot f^{int}(I,i,j) \cdot \Delta^{sat}(I,i,j)
\end{equation}
with $\Delta^{sat}$ denoting the saturation difference between the pixel $i$ and its $8$ neighbor $j$, see equation \eqref{eq:Delta3}:
\begin{equation}
\label{eq:Delta3}
\Delta^{sat}(I,i,j)=\vert Sat(I,i) - Sat(I,j) \vert
\end{equation}

5.\textit{contrast} \textit{of} \textit{intensity}: a contrast is visible when dark colors and shiny ones coexist. The bright colors
attract visual attention unless a dark region is completely surrounded by highly bright regions. The contrast of intensity is defined by equation \eqref{eq:IntensityContrast}:
\begin{equation}
\label{eq:IntensityContrast}
X_5 (I,i)= \sum_{j \in \eta_i}  f^{sat}(I,i,j) \cdot f^{int}(I,i,j) \cdot \Delta^{int}(I,i,j)
\end{equation}
With $\Delta^{int}$ denotes the difference of intensity between the pixel $i$ and its $8$ neighbor $j$
\begin{equation}
\Delta^{int}(I,i,j)=\vert Int(I,i) - Int(I,j) \vert
\end{equation}

6. \textit{dominance} \textit{of} \textit{warm} \textit{colors}: the warm colors -red, orange and yellow- are visually attractive.
These colors ($hue < 0.125 \simeq 45^\circ $) are still visually appealing, although the lack of contrast (hot and
cold colors in the area) is observed in the surroundings. This feature is defined by equation \eqref{eq:DominantContrast}:
\begin{eqnarray}
\label{eq:DominantContrast}
 V_6(I,i) = \begin{cases}
 Sat(I,i) \cdot Int(I,i)\ \ \ \ \ \ \ \ \ \ \ \ \     if\ 0 \leq Hue(I,i) < 0.125 \\
 0 \ \ \ \ \ \ \ \ \ \ \ \ \ \ \ \ \ \ \ \ \ \ \ \ \  \ \ \ \ \ \ \ \ \ \ \ \ \  otherwise
 \end{cases}
\end{eqnarray}

7. \textit{dominance} \textit{of} \textit{brightness} \textit{and} \textit{saturation}: highly bright, saturated colors are considered attractive
regardless of their hue value. The feature is defined by equation \eqref{eq:DominanceBrightness}:
\begin{equation}
\label{eq:DominanceBrightness}
V_7(I,i) = Sat(I,i) \cdot Int(I,i) 
\end{equation}
The normalization ($V_{1 \cdots 5} (I,i) = \frac{X_{1 \cdots 5}}{\vert\eta_i\vert}  $ ) of the first five descriptors ($X_{1 \cdots 5} $) by the
number of neighboring pixels ($\vert \eta_i \vert=8$) is performed.
In \cite{New32}, \cite{New17} it is reported that mixing a large quantity of different features increases the performance of prediction. This is why 
it is attractive to mix primary features (1-7) with those which have been used in previous works of saliency prediction \cite{New5}, that is simple RGB planes of a video frame. 



\subsection{The network design}
\label{Network}
\begin{figure}[H]
\begin{center}
\includegraphics[width=0.9\linewidth]{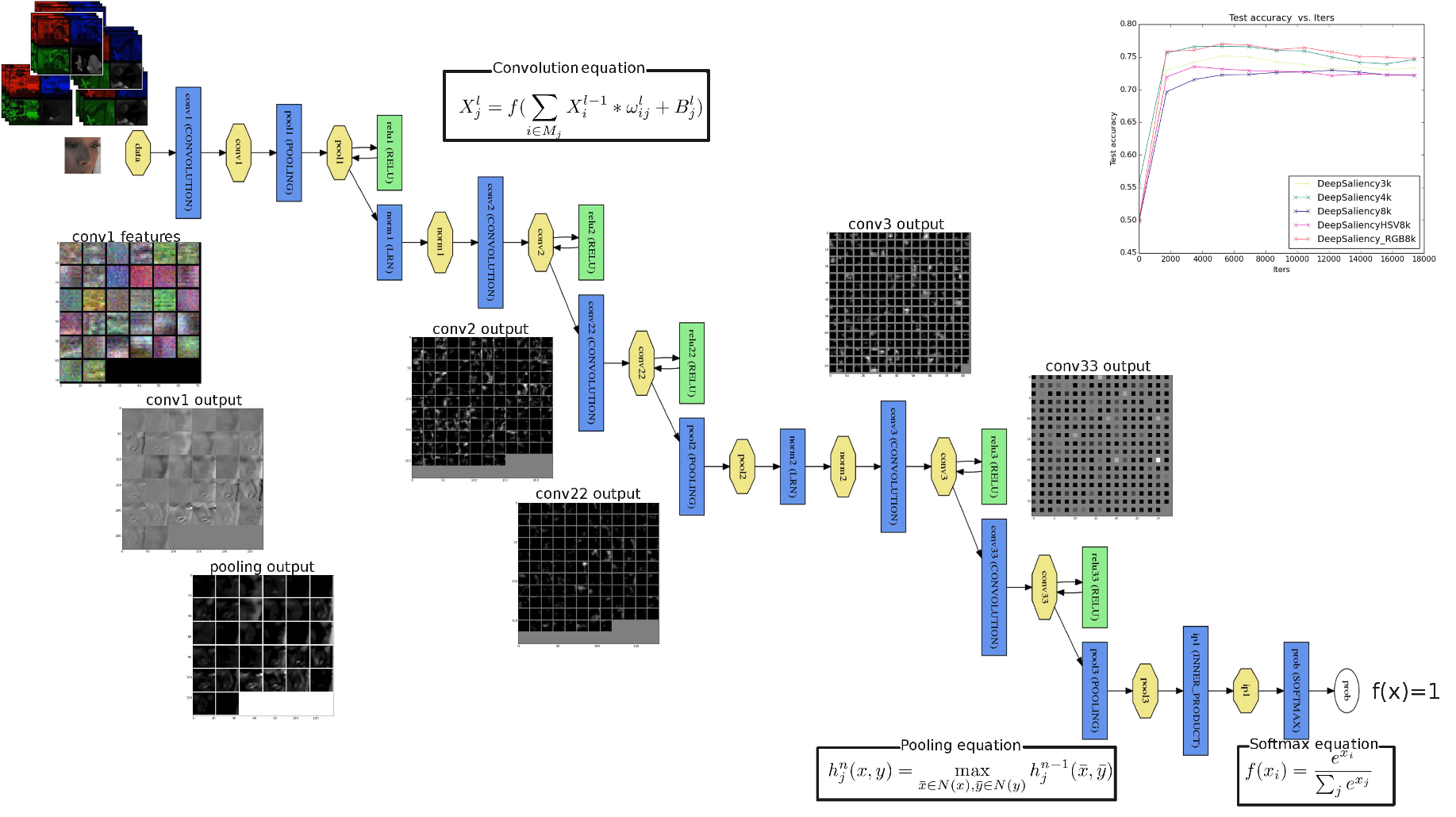}
\caption{\label{resumeDeepArchitecture} Architecture and design of the deep saliency framework.}
\end{center}
\end{figure}

\begin{figure}[H]

\begin{tabular}{ c c }
 &  \multirow{5}*{\includegraphics [width=0.25\textwidth]{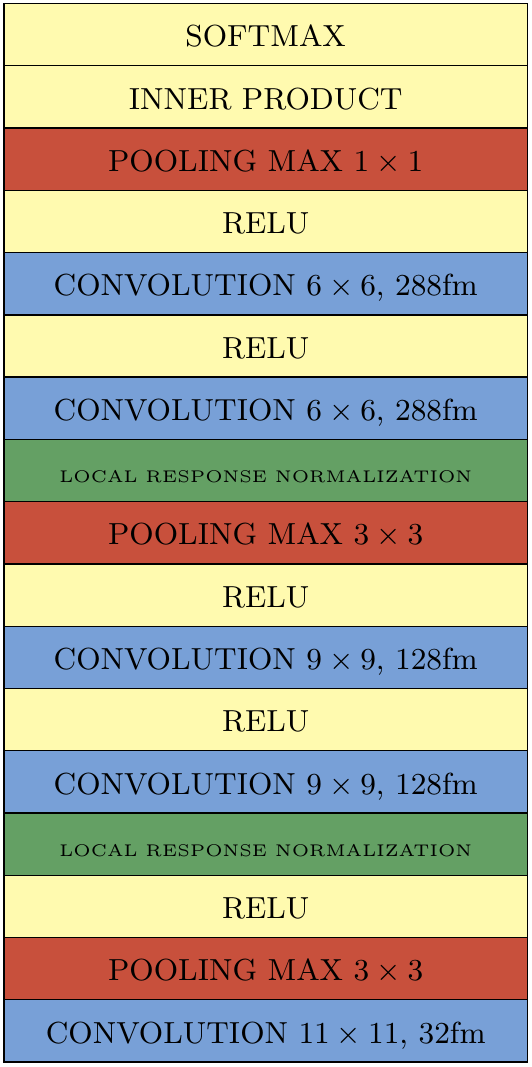}} \\
  & \\
  & \\
\includegraphics [width=0.7\textwidth]{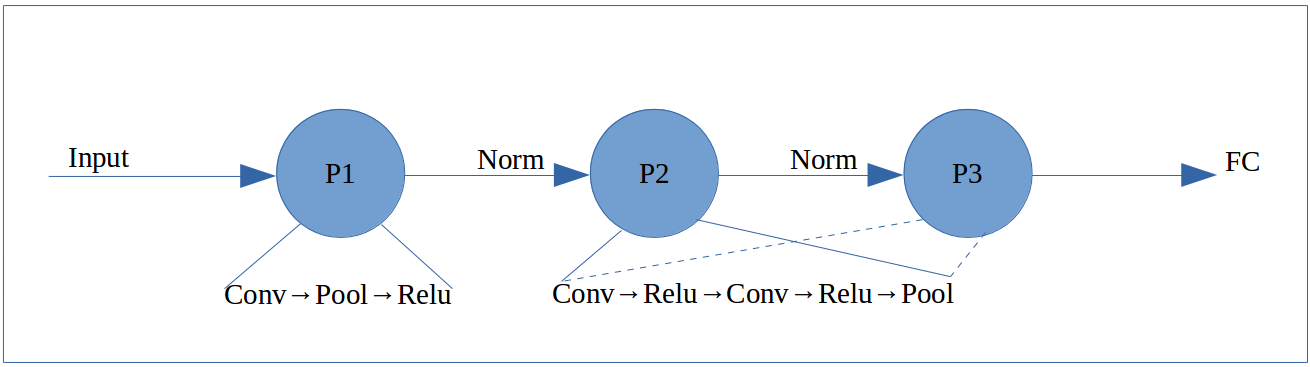} & \\
 & \\
 & \\

\end{tabular}
\caption{\label{networkArchitec}Architecture of video saliency convolution network}
\end{figure}

In this section we present the architecture of a deep CNN we designed for our two class classification problem: prediction of a saliency of a patch in a given video frame.  
It includes five layers of convolution, three layers of pooling,  
five layers of Rectified Linear Units (RELU), two normalisation layers, 
and one layer of Inner product followed by a loss layer as illustrated in Figure \ref{resumeDeepArchitecture}. 
The final classification is ensured by a soft-max classifier in equation \eqref{eq:Softamx}. This function is a generalization of the logistic 
function that compresses a vector of arbitrary real values
of $K$ dimension to a vector of the same size but with actual values in the range
$(0, 1)$.
\begin{equation}
\label{eq:Softamx}
 f(x_i)=\frac{e^{x_i}}{\sum_j e^{x_j}}
\end{equation}
Figure \ref{networkArchitec} shows  the order of layers in our  proposed network. The CNN architecture was implemented using the Caffe software \cite{New13}.
We created our network architecture made on three patterns (see figure \ref{networkArchitec} with a step of normalisation between each one. Each pattern contains a linear/nonlinear cascading operation (convolution, pooling, RELU). 
For the first pattern we chose a cascading operation different than the two following patterns. The first operation cascade is represented as the succession of convolution layer,
pooling layer followed by a RELU layer. In fact, the applying of the pooling operation before the RELU layer does not change the final results because the two layers compute 
the function of maximum, however, it ensures the decrease of the execution time of the prediction as the step of pooling reduces the number of nodes.
The two convolution layers stacked before the pooling layer for the followed pattern ensures the  development of more complex features that will be more "expressive" before the destructive Pool
operation.
In the following, we will describe the most crucial layers which are convolution, pooling and local response normalisation.
\subsubsection{Convolution layers } In order to extract the most important information for further analysis or exploitation
of image patches, the convolution  with a fixed number
of filters that is based on the natural functioning of the HVS is needed.
It is necessary to determine the size of the convolution kernel to be applied to each pixel of the input image to highlight areas of the image.
Gaussian filters were used to create all of the feature maps of the convolution layer. The number of filters, in other words the number of kernels, 
convolved with the input image is the number of the obtained feature maps. Three stages are conceptually necessary to create the convolution layer. 
The first refers to the convolution of the input image by linear filters.
The second is to add a bias term. And finally, the application of a nonlinear function (here we have used the rectified linear function $f(x)=max(x,0)$).
Generally, the equation of convolution can be written as\eqref{eq:Convolution}:
\begin{equation}
\label{eq:Convolution}
 X_j^l=f( \sum_{i \in M_j} X_{i}^{l-1} * \omega_{ij}^l +B_j^l )
\end{equation}
with $X_j^l$ : the activity of the unit $j$  according to the layer $l$,

$X_i$ represents a selection of the input feature maps,

$B_j^l$ is the additive bias of the unit $j$ in the features maps of the layer $l$,

$\omega_{ij}^l$: presents the synaptic weights between unit $j$ of the layer $l$ and $l-1$.

\subsubsection{Pooling layers } 
To reduce the computational complexity for the upper layers, and provide a form of translation invariance, pooling summarizes the outputs of neighboring groups of neurons on the same 
kernel map.
The size of the region of 'pooling' reduces the size of each feature map as input by the acquisition of a value for each region. We use max-pooling, see equation \eqref{eq:MaxPooling}:  
\begin{equation}
\label{eq:MaxPooling}
 h_{j}^{n}(x,y)=\max _{\bar{x}, \bar{y} \in \mathscr{N} }   h_{j}^{n-1}(\bar{x},\bar{y})
\end{equation}
Here $\mathscr{N}$ denotes the neighbourhood of (x,y). 
\subsubsection{Local response normalization layers } LRN layer normalizes values of feature maps which are calculated through the neurons having unbounded activations to detect the high-frequency 
characteristics with a high response of the neuron, and to amortize answers that are uniformly greater in a local area. The output 
computation is presented is presented in equation \ref{eq:LocalResponseNormalization}:
 \begin{equation}
 \label{eq:LocalResponseNormalization}
  f(U_{f}^{x,y})=\frac{U_{f}^{x,y}}{(1+\frac{\alpha}{N^2} \sum_{x'= \max(0,x-[N/2])}^{\min(S, x-[N/2]+N)}  \sum_{y'= \max(0,y-[N/2])}^{\min(S, y-[N/2]+N)} 
  (U_{f}^{x',y'})^2)^\beta}
 \end{equation}
 
 Here $U_{f}^{x,y}$ represents the value of  the feature map at $(x,y)$ coordinates and the sums are taken in the neighbourhood of $(x,y)$ of size $N\times N$,
 $\alpha$ and $\beta$ regulate normalisation strength. 
\subsection{Training and validation of the model} 
To solve the learning problem and to validate the network with the purpose to generate a robust model of salient area recognition, 
the solver of Caffe \cite{New13} is iteratively optimizing the network parameters in forward-backward loop.
The optimisation method used is that one of stochastic gradient. 
The parameterization  of the solver requires setting the learning rate and the number of iterations at  training and testing steps. 

The numbers of training and testing iterations are defined according to the "batch size" parameter of Caffe \cite{New13}. 
The batch size presents the number of images that is salient and non-salient patches in our case, processed at an iteration.
This number depends on two parameters:
\begin{itemize}
 \item The power of the GPU/RAM of the used machine, 
 \item The number of patches available for each database.
\end{itemize}

The number of iterations is computed according to equation \eqref{eq:IterationNumbers}:
\begin{equation}
\label{eq:IterationNumbers}
  iterations\_numbers = epochs  \times  \frac{Total\_images\_number}{batch\_size} 
\end{equation}
here $batch\_size$ represents the number of images for each network switching, $epochs$ presents how many times 
the totality of the dataset is switched by the network.

\begin{figure}[H]
\begin{center}
\renewcommand{\arraystretch}{2}
   \begin{tabular}{ | p{1cm} p{4cm} p{4cm} |}
   \hline
        & \multirow{3}{*}{\includegraphics[width=1\linewidth]{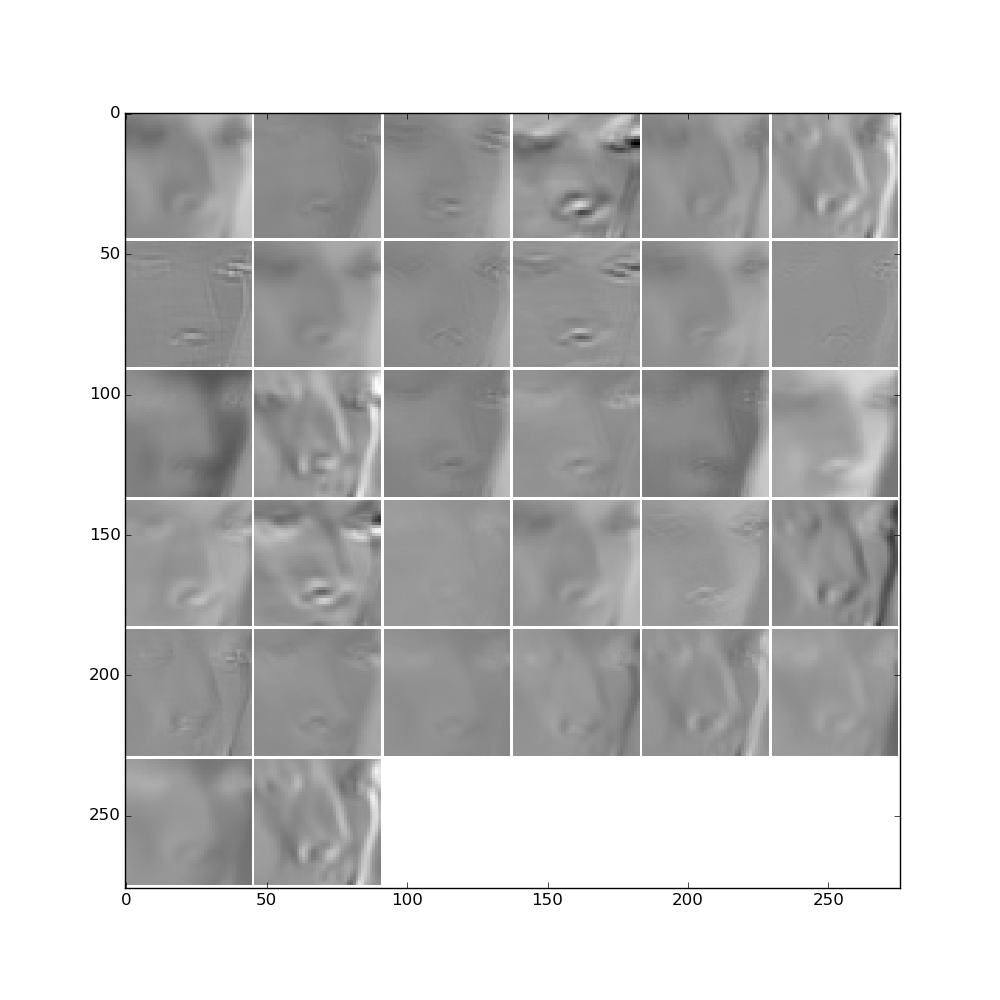}} & \multirow{3}{*}{\includegraphics[width=1\linewidth]{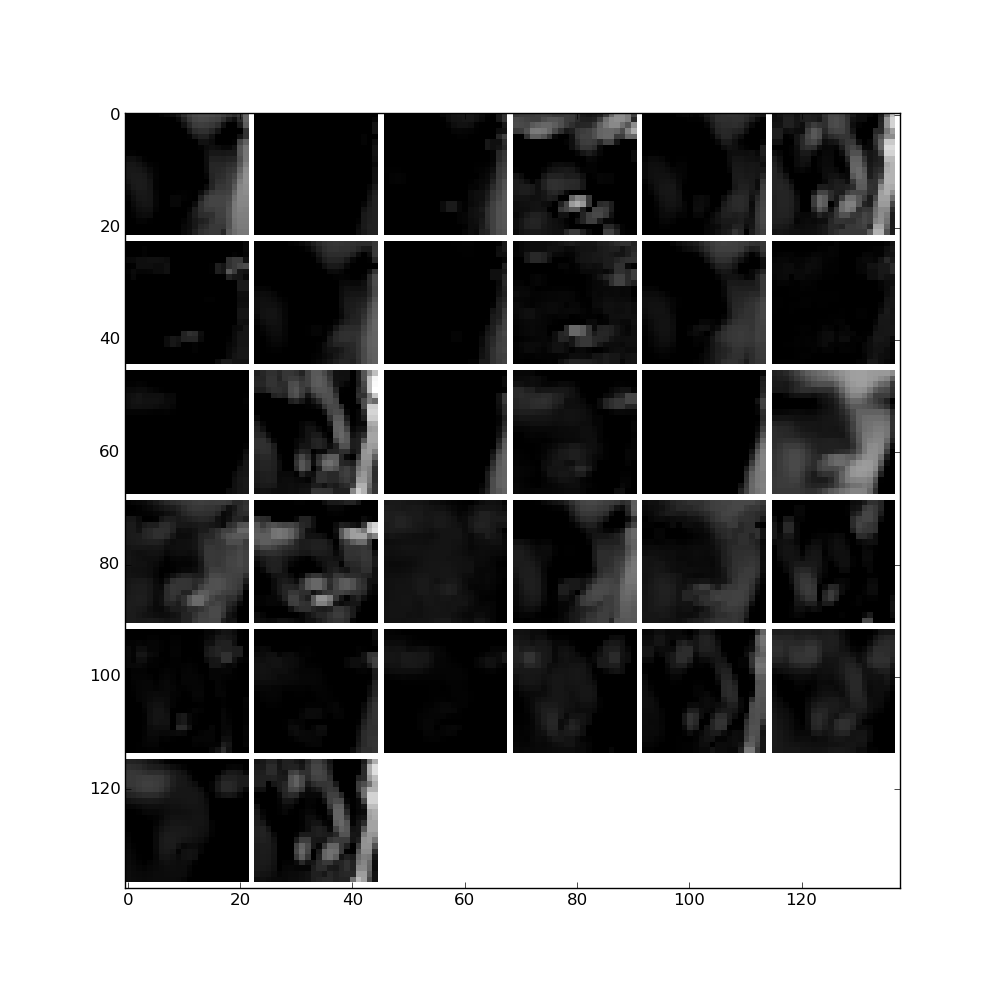}}\\ 
        \includegraphics[width=1\linewidth]{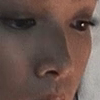} &  &\\
        $\ \ $ \tiny{(a)} & $\ \ \ \ \ \ \ \ \ \ \ \ \ \ \ \ \  \ \ \ \ \ \ \ \  \ \ \ \ \ \ \ \ \ \ \ \ $\tiny{(b)} & $\ \ \ \ \ \ \ \ \ \ \ \ \ \ \ \ \ \  \ \ \ \ \ \ \ \ \ \ \ \ \ \  \ \ \  \ \ \ $\tiny{(c)}\\
   \hline
   \end{tabular}
\end{center}
   \caption{\label{conv1_output}(a) Input patch, (b) the output of first convolution layer and (c) the output of the first pooling layer.}
\end{figure}

\begin{figure}[H]
\begin{center}
\fbox{\rule{0pt}{1in} 
   \includegraphics[width=0.37\linewidth]{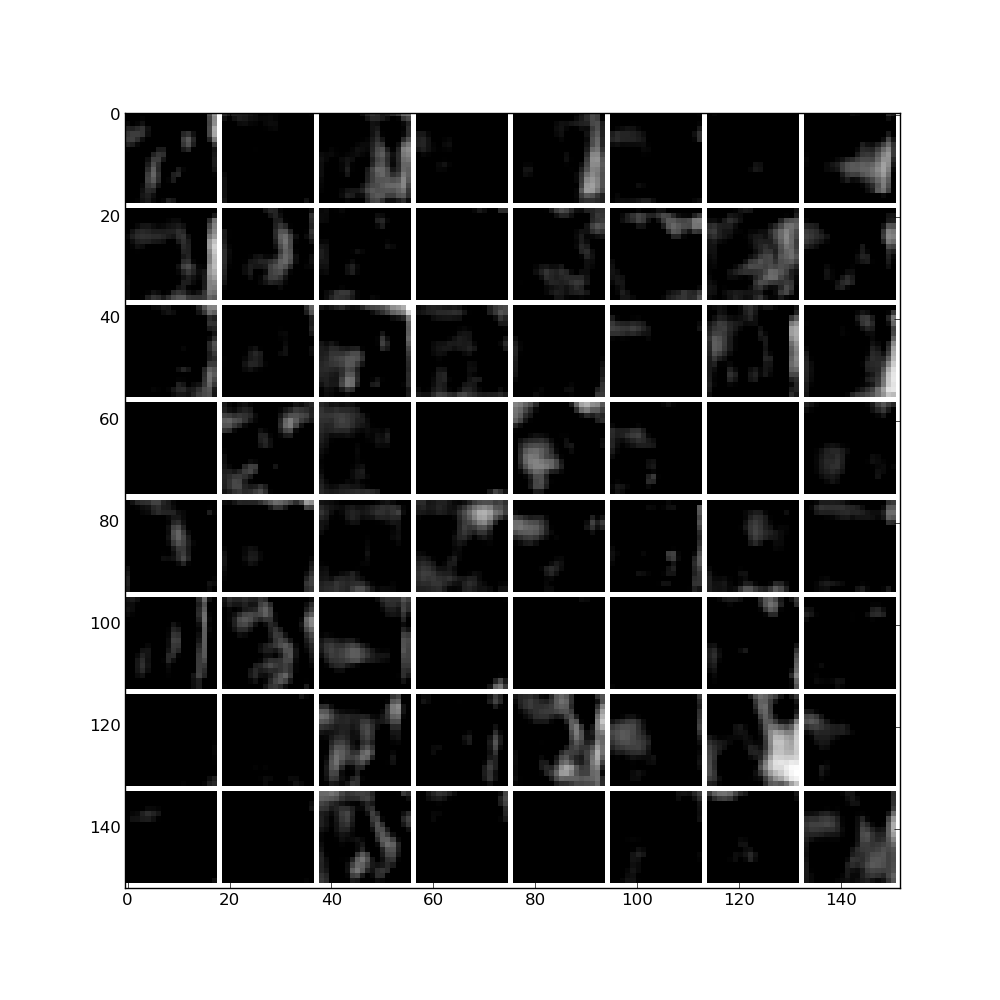}
   \includegraphics[width=0.37\linewidth]{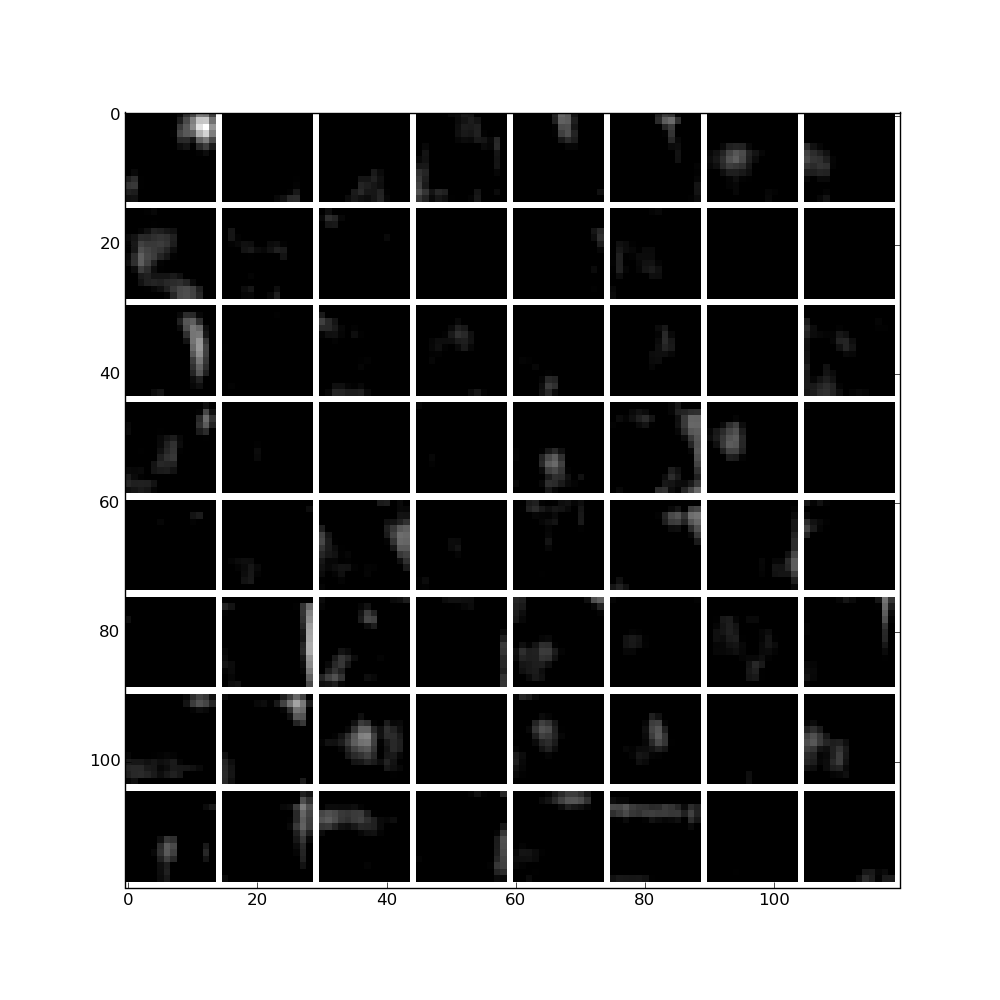}
   \rule{0.01\linewidth}{0pt}}
\end{center}
   \caption{\label{conv2_output}The output of the 2nd convolution layer data of ' Conv2' and 'Conv22' .}
\label{fig:long}
\label{fig:onecol}
\end{figure}

\begin{figure}[H]
\begin{center}
\fbox{\rule{0pt}{1in} 
   \includegraphics[width=0.37\linewidth]{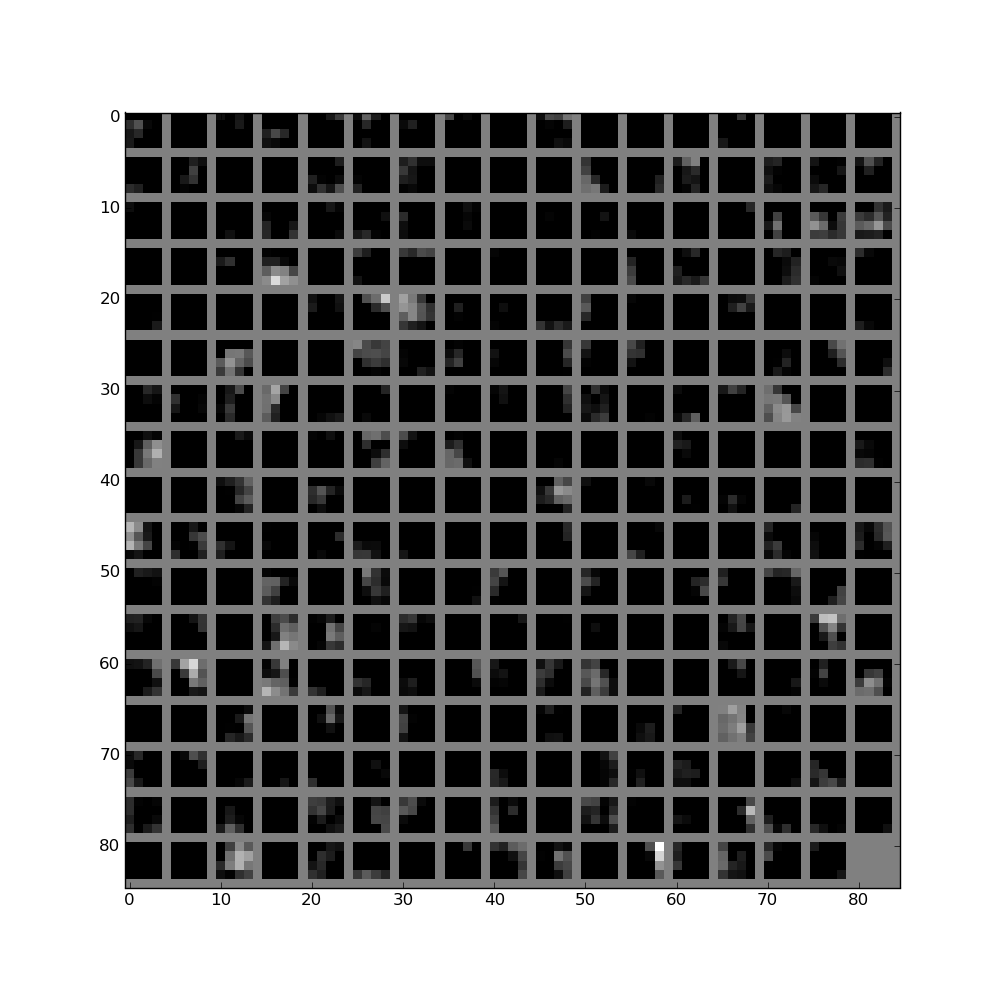}
   \includegraphics[width=0.37\linewidth]{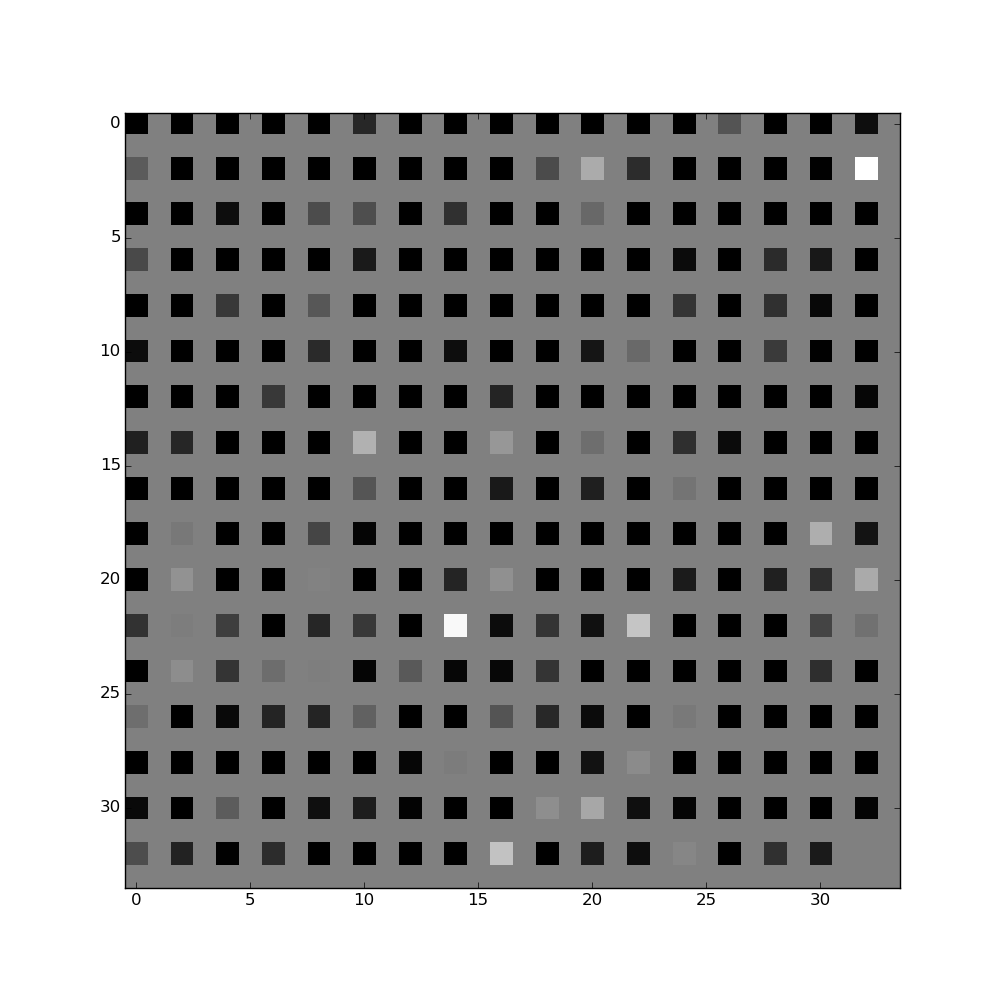}
   \rule{0.01\linewidth}{0pt}}
\end{center}
   \caption{\label{conv3_output}The output of third convolution layer ' Conv3' and 'Conv33'.}
\label{fig:long}
\label{fig:onecol}
\end{figure}

It is interesting to visualize the purely spatial features computed by the designed CNN in case when the network is configured
to predict saliency only with primary RGB values  as this it the goal instead of aspiration of the overall deep learning approach to saliency prediction. 
As the feature integration theory states, the HVS is sensitive to orientations and contrasts. This is what we observe in features going 
through layers of the network.  The output of convolution layers (see figures \ref{conv1_output}, \ref{conv2_output}
and \ref{conv3_output}) yields more and more contrasted and structured patterns. In this figure $convi$ and $convii$ stands for consecutive 
convolution layers without pooling layers in between. 

\subsection{Generation of a pixel-wise saliency map}

The designed and trained Deep CNN predicts for a patch in a video frame if it is  salient for a human observer.
Despite the interest of this problem for selection of important areas in images for further pattern recognition tasks, for finer, pixel-wise saliency prediction in video, the transformation of sparse classifier responses into a dense predicted saliency map is needed. 
The response for each patch is given by the soft-max classifier, see figure \ref{resumeDeepArchitecture} and equation \eqref{eq:Softamx} in section \ref{Network}. 
The value of classifier which is interpreted as a probability to belong to the saliency class,  can be considered as a predicted saliency of a patch. 
Then a Gaussian is centred on the patch center with a pick value of $\frac{10 f(i)}{2 \pi \sigma^2} $ with the spread parameter $\sigma$ chosen of a half-size of the patch.
Hence a sparse saliency map is predicted. 
In order to densify the map we classify densely sampled patches with a half-patch overlap and then interpolate obtained values.  
Examples of predicted saliency maps using RGB only features (3K model), RGB features and Residual motion features(4Kmodel),
Wooding gaze-fixation maps and popular saliency prediction models of Itti \cite{New8}(named "GBVS") and Harell\cite{New15}(named "SignatureSal")
are depicted in table \ref{tab_wooding}. 
Visual evaluation of the maps shows that the proposed method yields maps more similar to Wooding maps built on gaze fixations. Indeed GBVS and SignatureSal are pixel-wise maps, while our maps are built upon salient patches. Further evaluation will be presented in the next section \ref{ExperimentsAndResults}. 

\begin{table}[H]
\caption{\label{tab_wooding} Different saliency map of testing frame from $576i50$ videos of IRCCYN database.}
 \begin{tabular}{ |c|c|c|c|c|c|c|}
 \hline
 \tiny{$nbr\_frame$} & \tiny{ Frame }& \tiny{ Wooding} & \tiny{Deep3k} & \tiny{Deep4k} & \tiny{GBVS} & \tiny{SignatureSal} \\ 
  \hline
  \tiny{$\#frame 34 $ }& \includegraphics[width=0.1\linewidth]{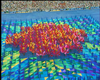} & 
  \includegraphics[width=0.1\linewidth]{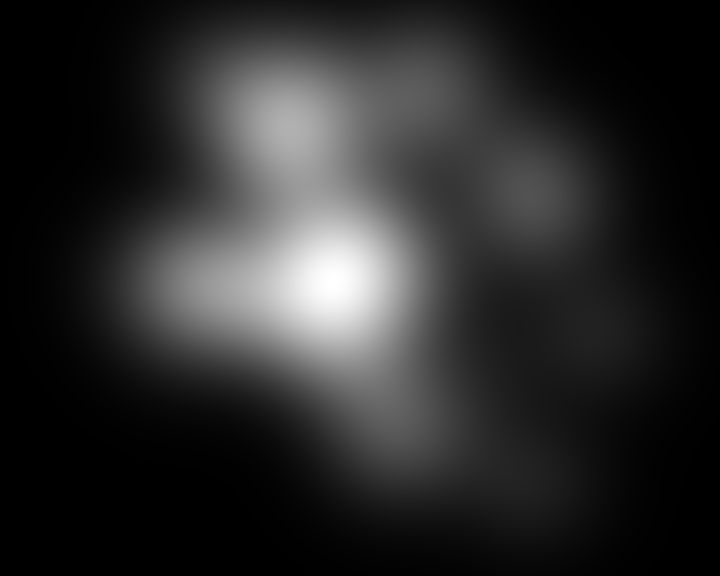} & 
  \includegraphics[width=0.1\linewidth]{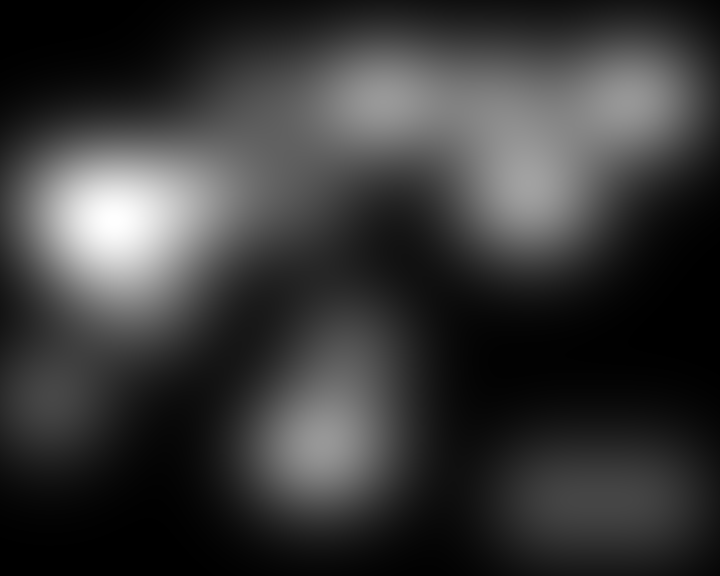} & 
  \includegraphics[width=0.1\linewidth]{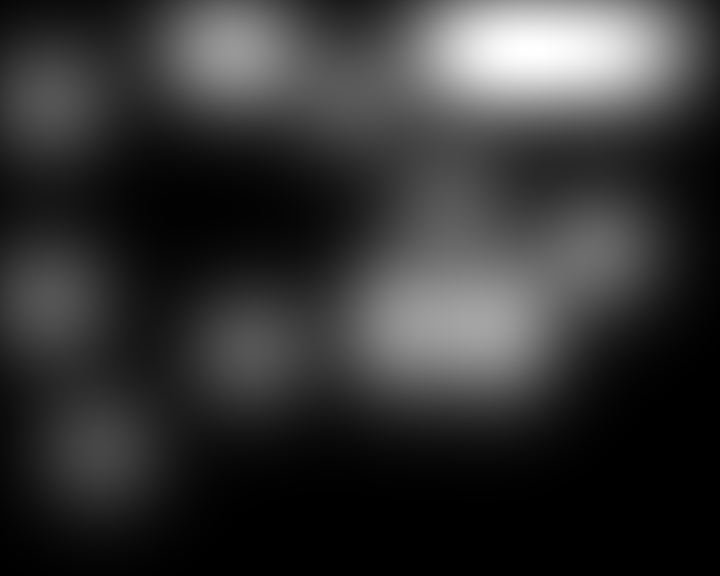} & 
  \includegraphics[width=0.1\linewidth]{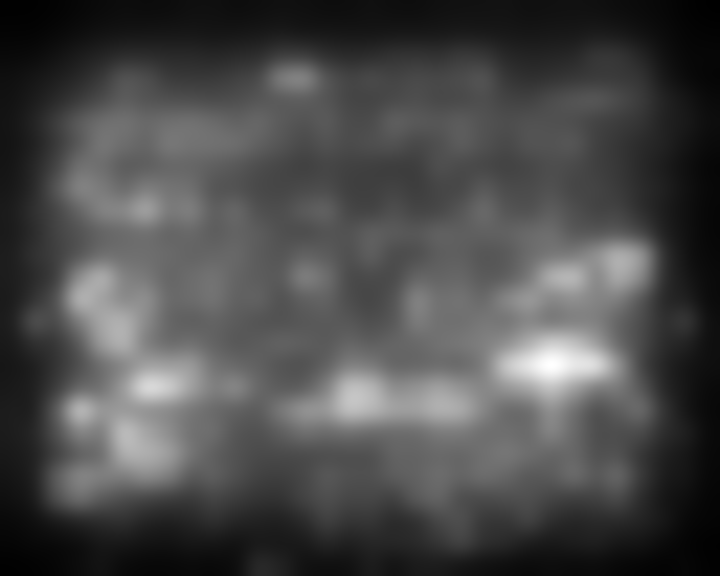} & 
  \includegraphics[width=0.1\linewidth]{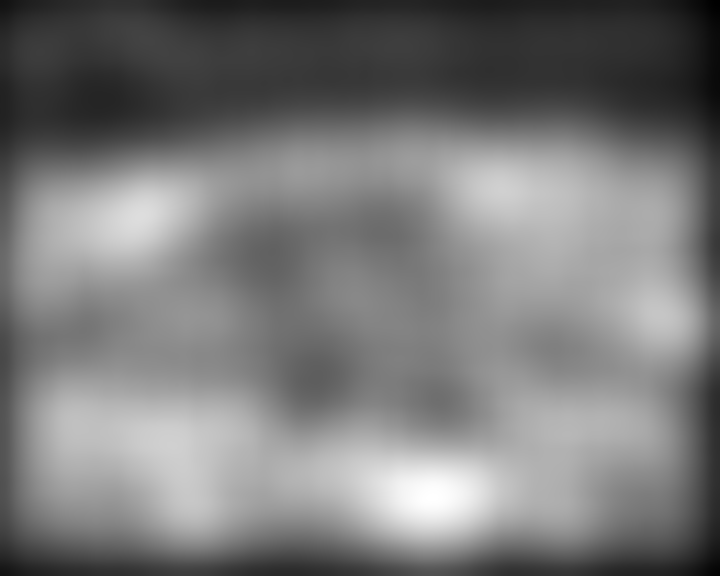} \\
  \hline
  \tiny{$\#frame 78$} & \includegraphics[width=0.1\linewidth]{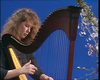} & 
  \includegraphics[width=0.1\linewidth]{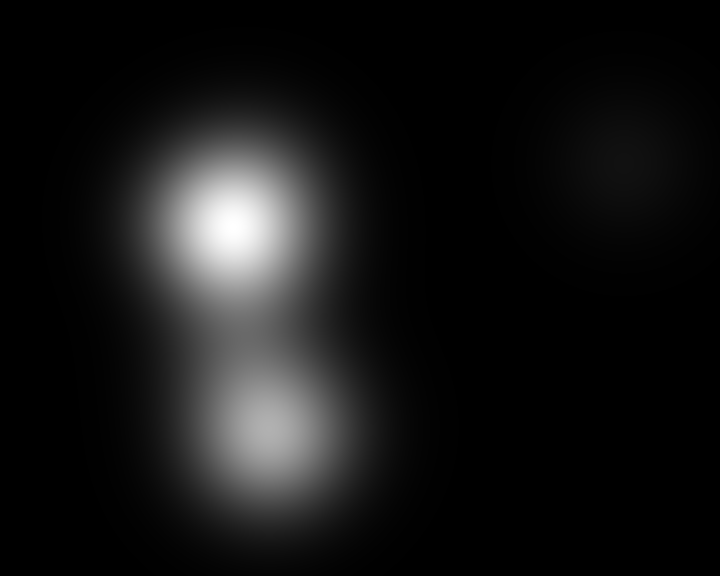} & 
  \includegraphics[width=0.1\linewidth]{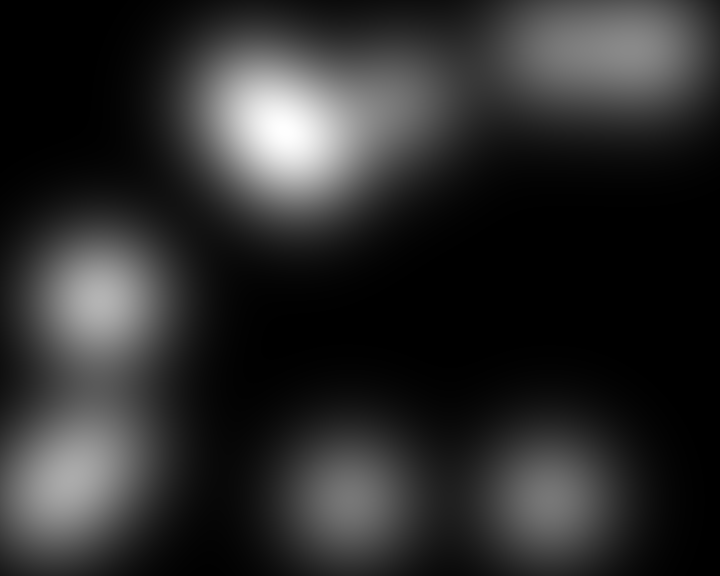} & 
  \includegraphics[width=0.1\linewidth]{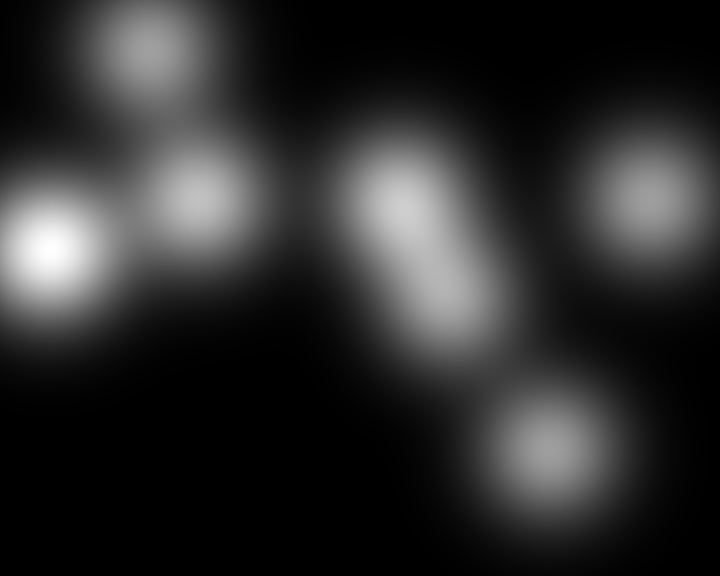} & 
  \includegraphics[width=0.1\linewidth]{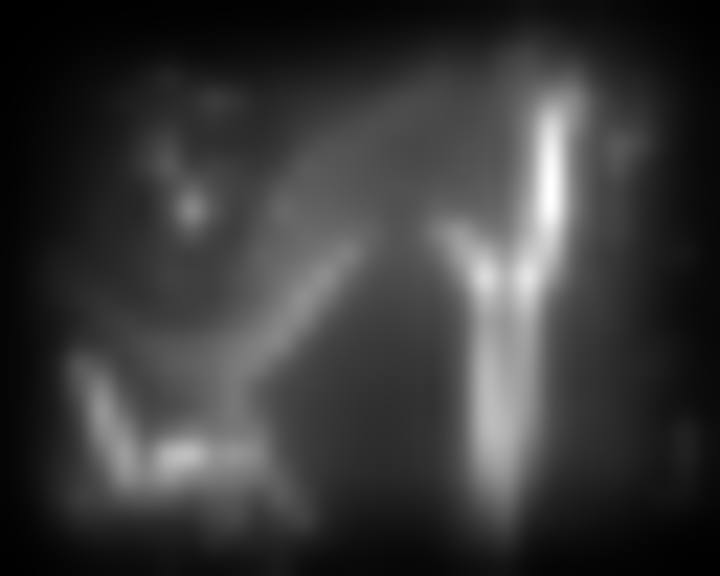} & 
  \includegraphics[width=0.1\linewidth]{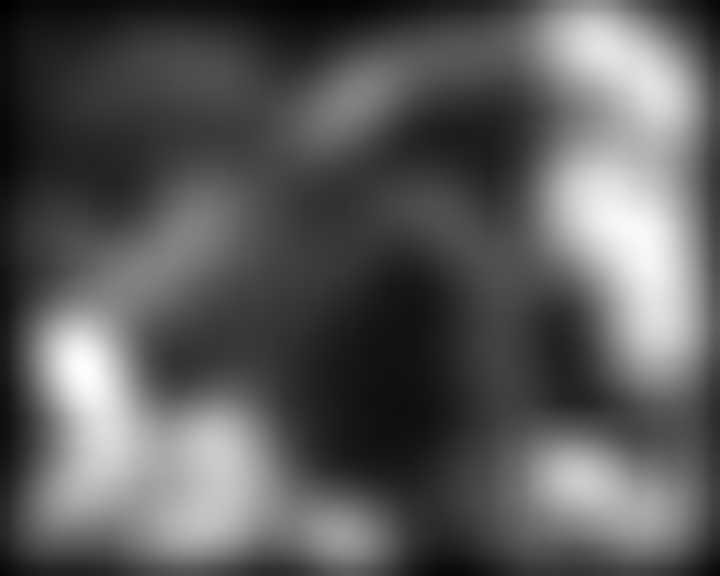} \\
  \hline
    \tiny{ $\#frame 122 $} & \includegraphics[width=0.1\linewidth]{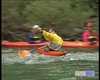} & 
  \includegraphics[width=0.1\linewidth]{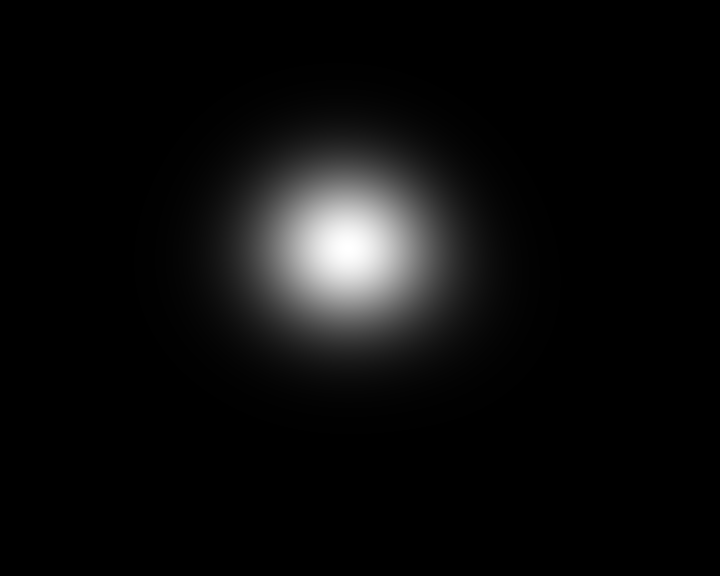} & 
  \includegraphics[width=0.1\linewidth]{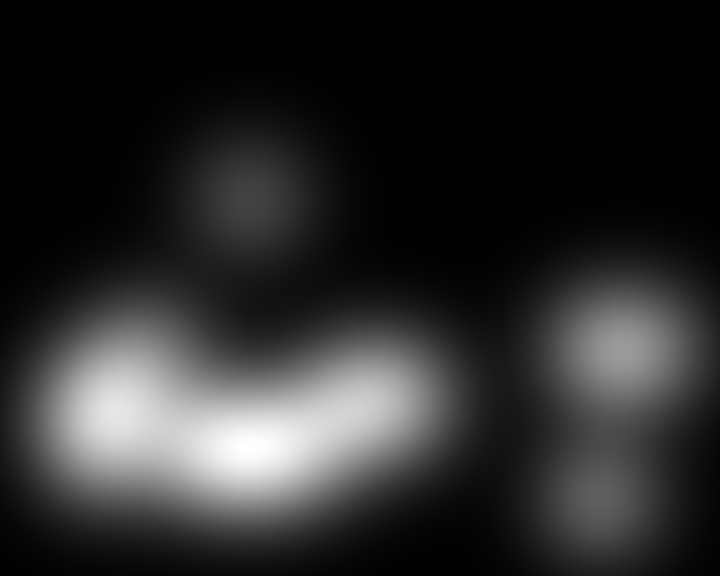} & 
  \includegraphics[width=0.1\linewidth]{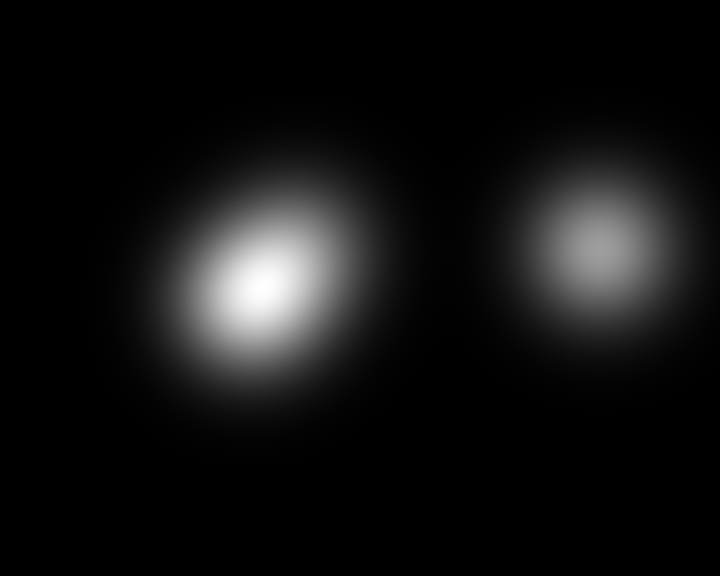} & 
  \includegraphics[width=0.1\linewidth]{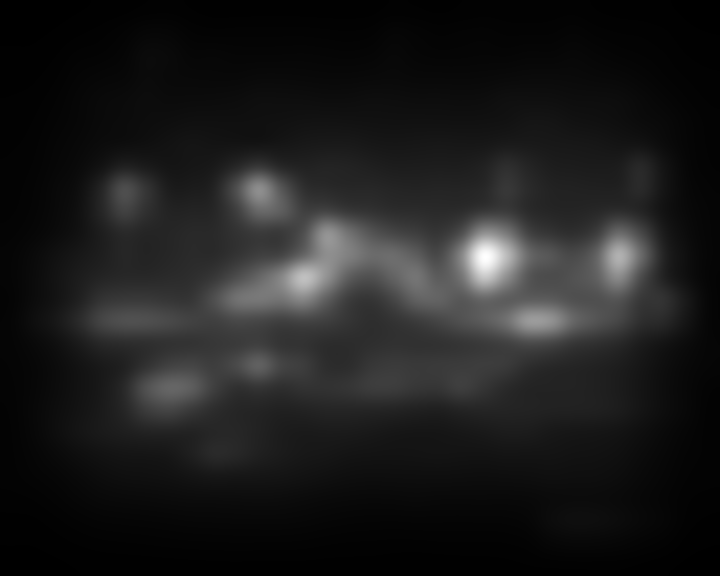} & 
  \includegraphics[width=0.1\linewidth]{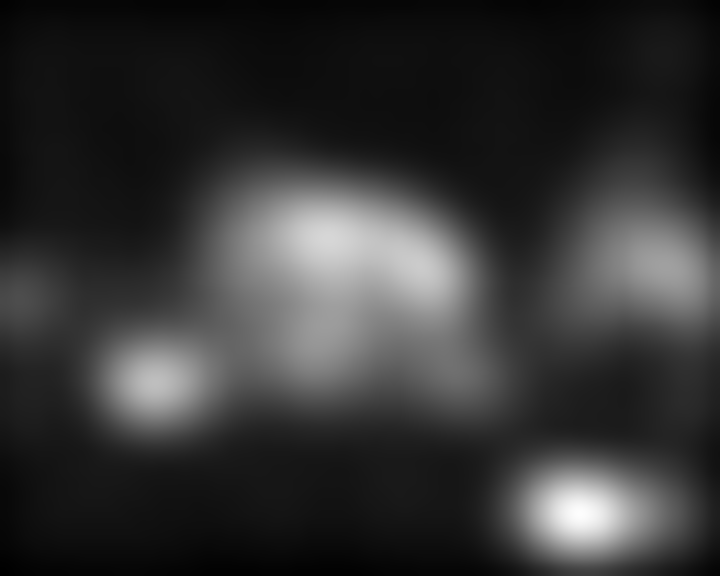} \\
  \hline
   \tiny{  $\#frame 166 $ }& \includegraphics[width=0.1\linewidth]{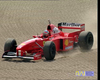} & 
  \includegraphics[width=0.1\linewidth]{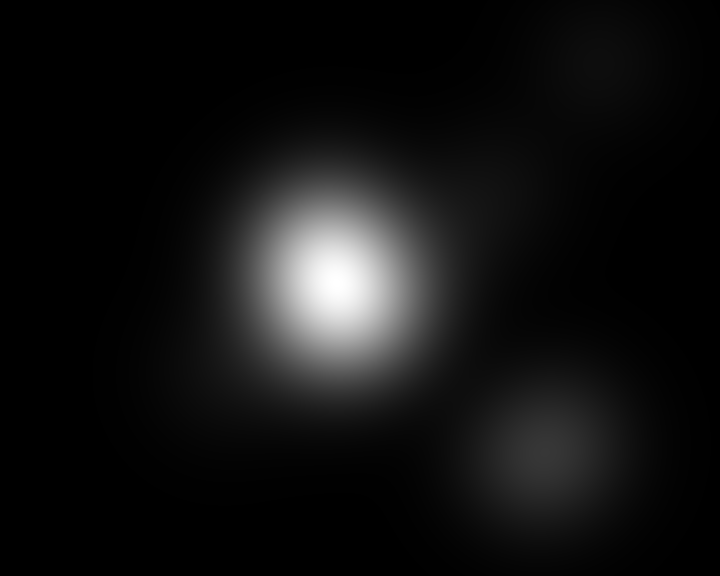} & 
  \includegraphics[width=0.1\linewidth]{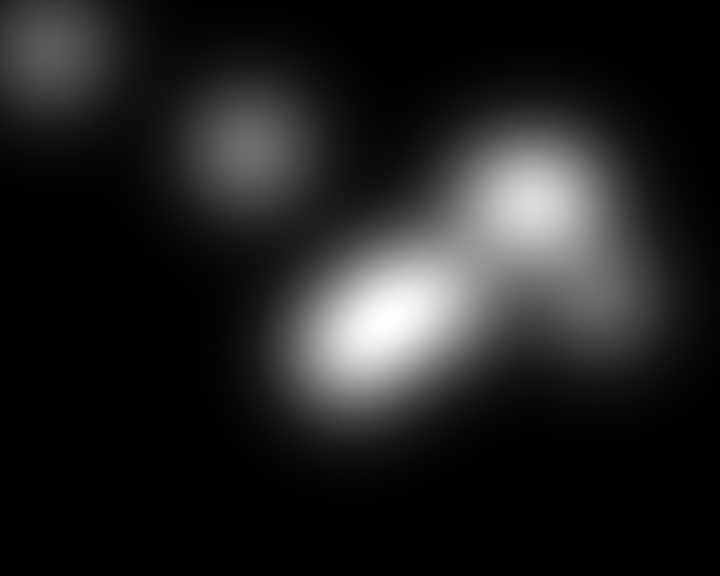} & 
  \includegraphics[width=0.1\linewidth]{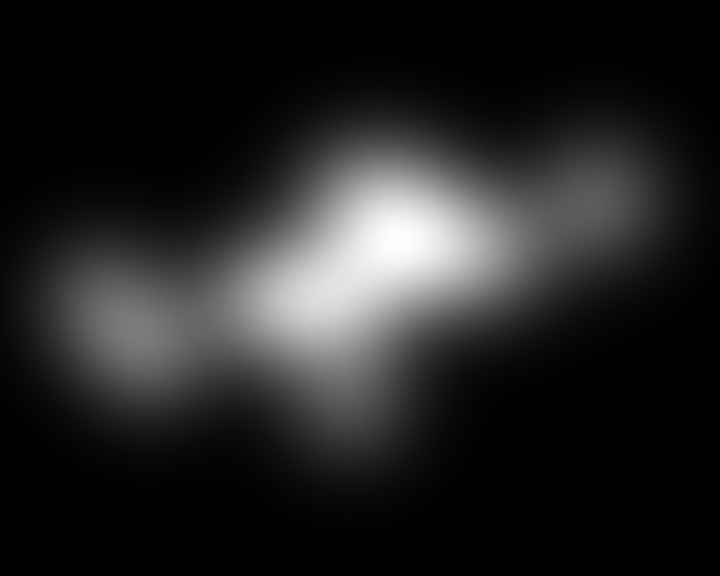} & 
  \includegraphics[width=0.1\linewidth]{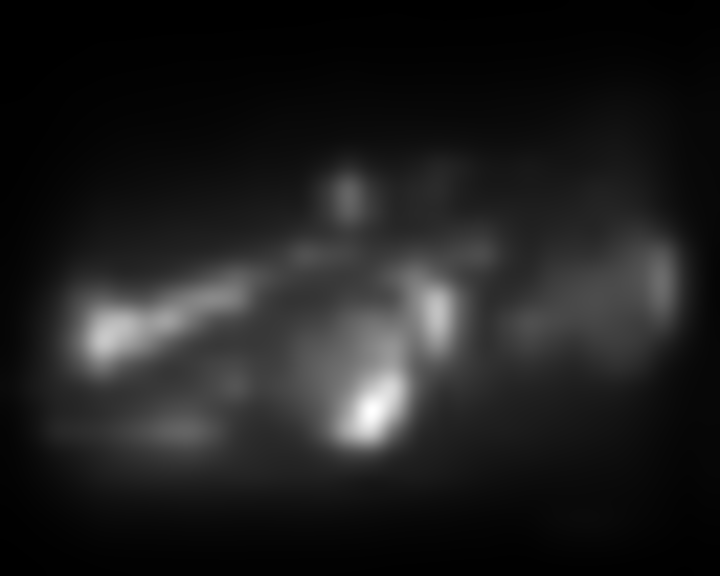} & 
  \includegraphics[width=0.1\linewidth]{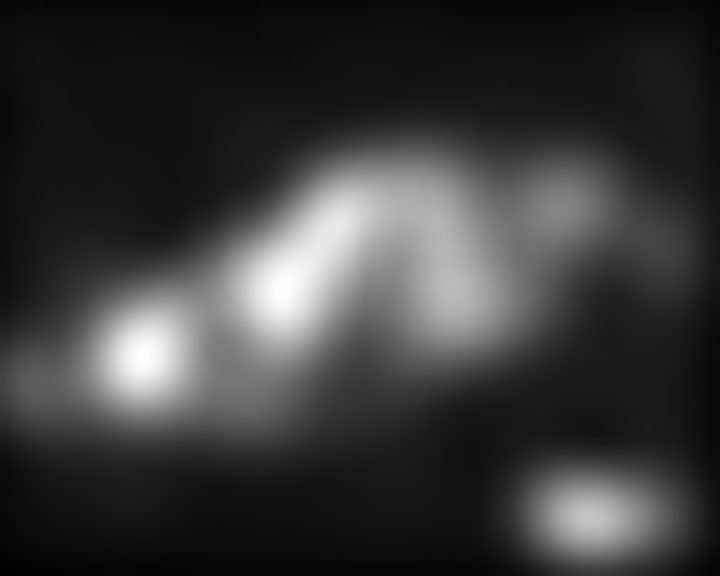} \\
  \hline
   \tiny{  $\#frame 217 $} & \includegraphics[width=0.1\linewidth]{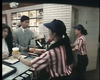} & 
  \includegraphics[width=0.1\linewidth]{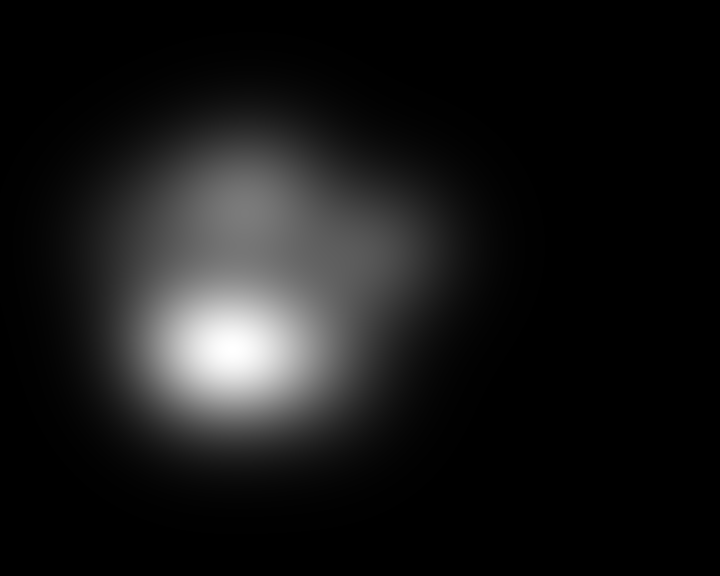} & 
  \includegraphics[width=0.1\linewidth]{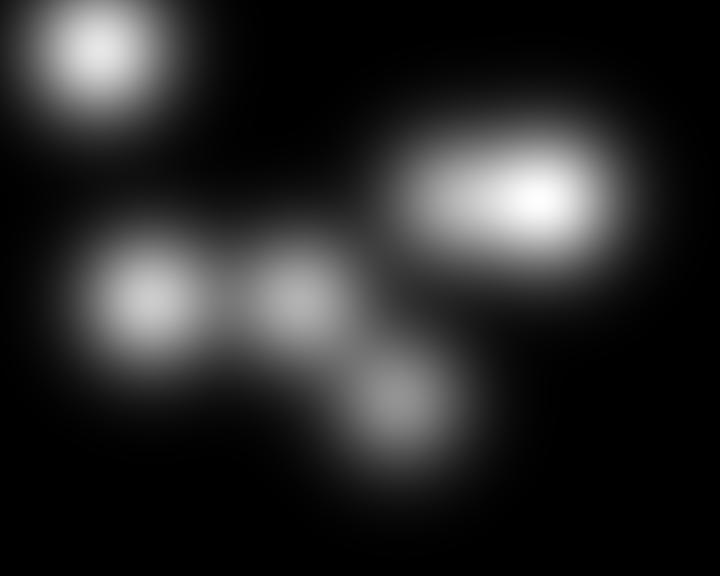} & 
  \includegraphics[width=0.1\linewidth]{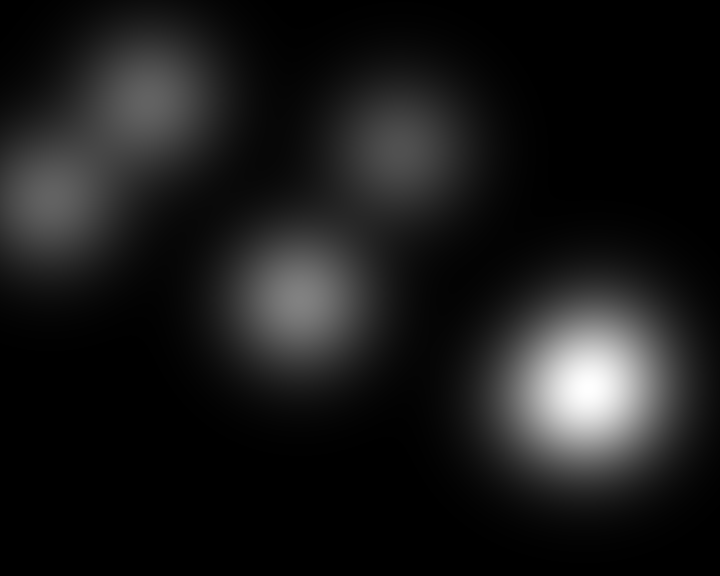} & 
  \includegraphics[width=0.1\linewidth]{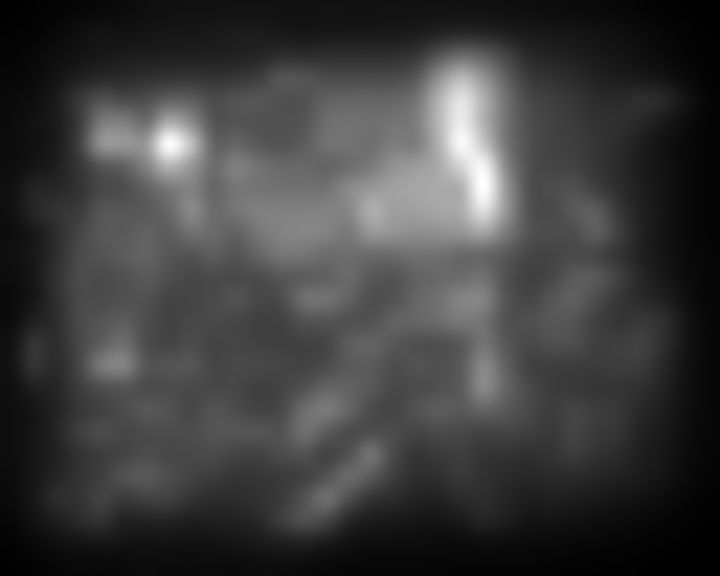} & 
  \includegraphics[width=0.1\linewidth]{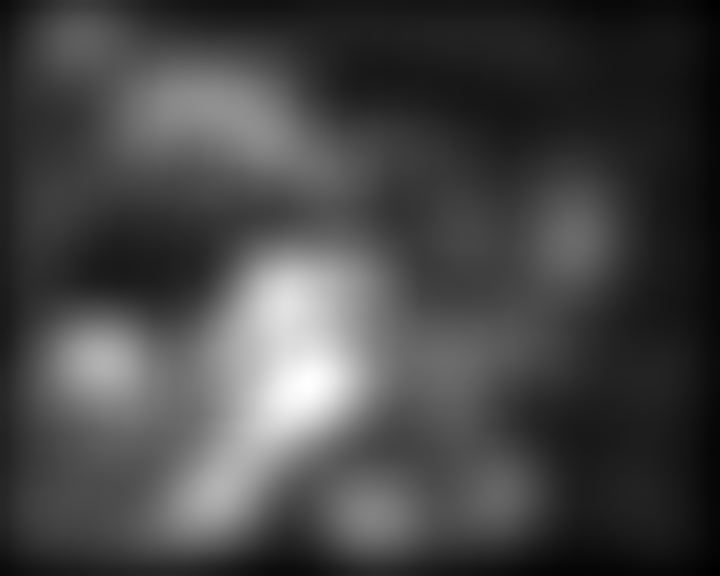} \\
  \hline
   \tiny{  $\#frame 253 $} & \includegraphics[width=0.1\linewidth]{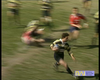} & 
  \includegraphics[width=0.1\linewidth]{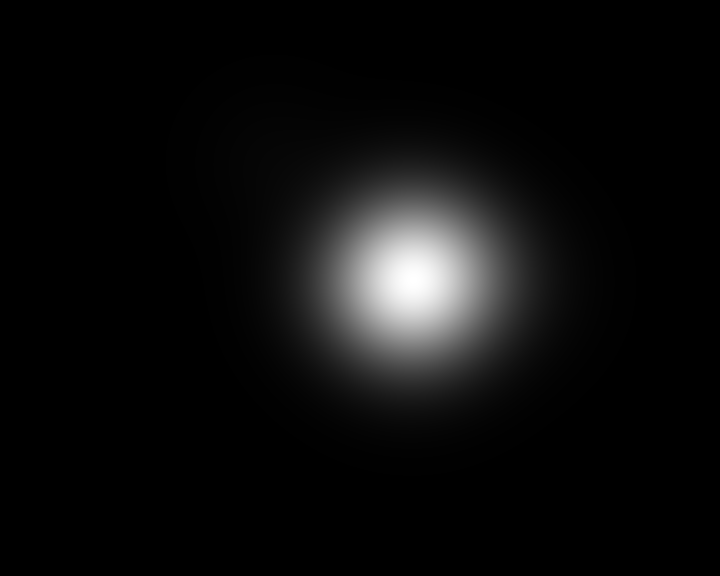} & 
  \includegraphics[width=0.1\linewidth]{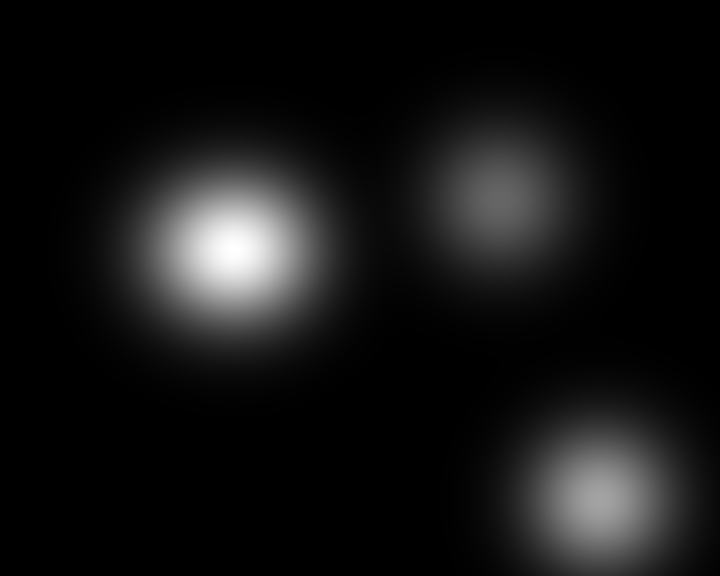} & 
  \includegraphics[width=0.1\linewidth]{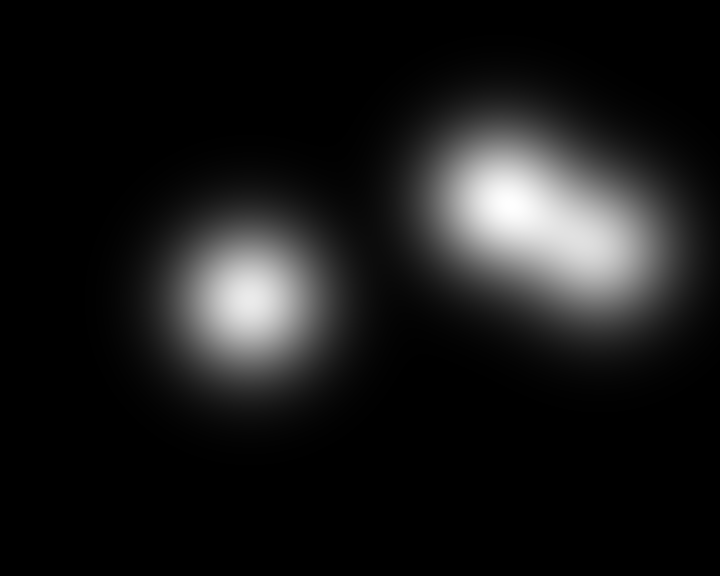} & 
  \includegraphics[width=0.1\linewidth]{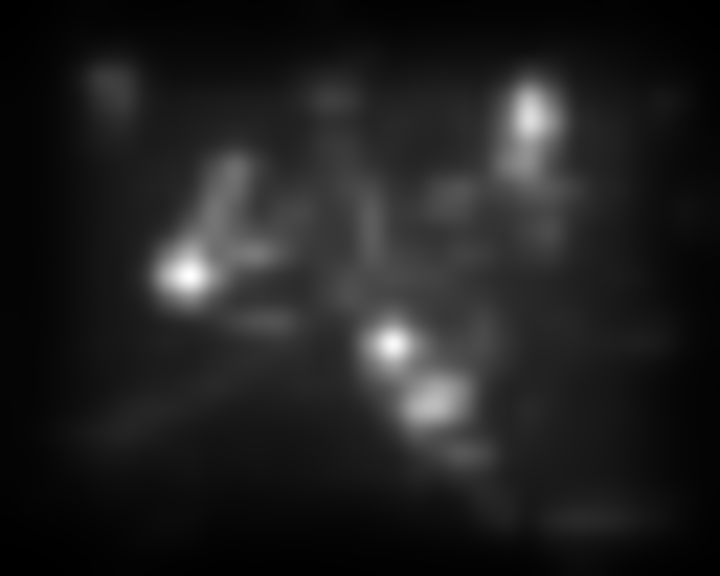} & 
  \includegraphics[width=0.1\linewidth]{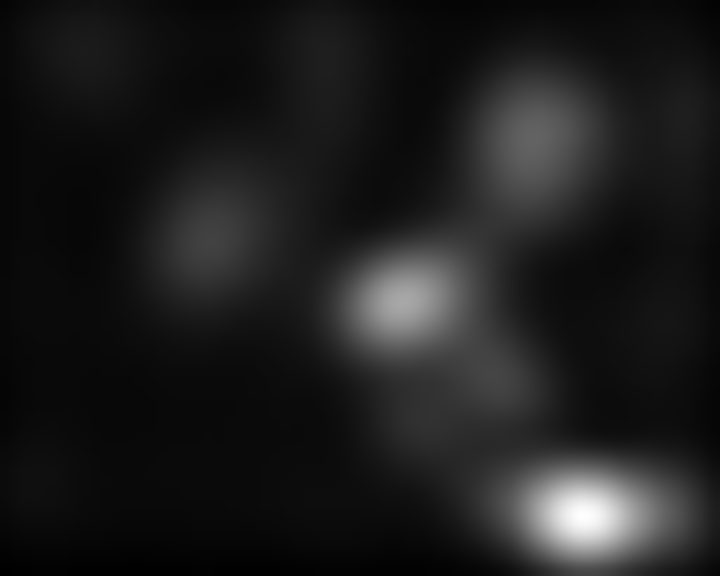} \\
  \hline
   \tiny{ $\#frame 298 $ }& \includegraphics[width=0.1\linewidth]{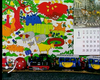} & 
  \includegraphics[width=0.1\linewidth]{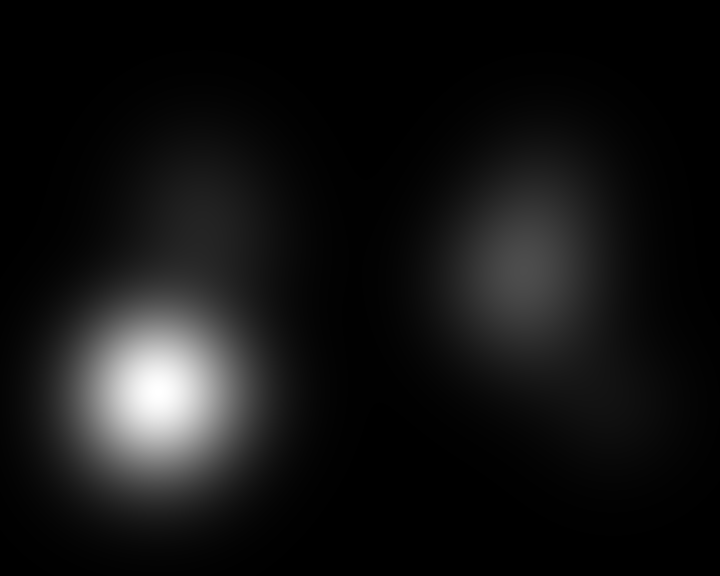} & 
  \includegraphics[width=0.1\linewidth]{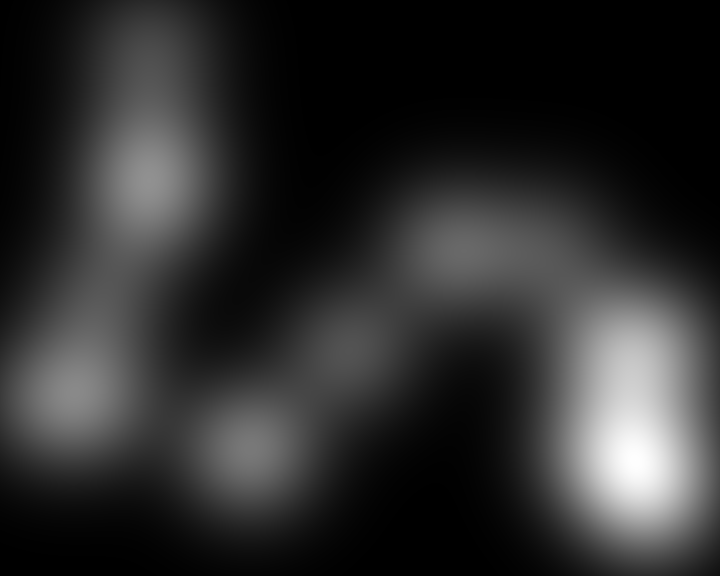} & 
  \includegraphics[width=0.1\linewidth]{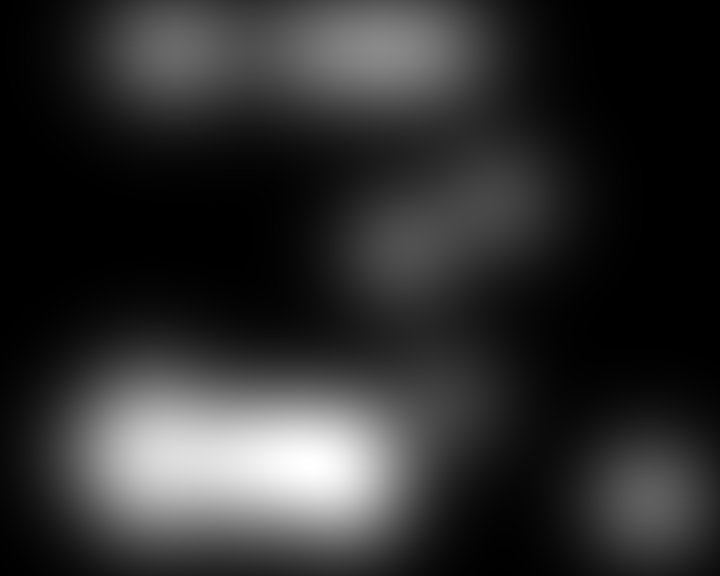} & 
  \includegraphics[width=0.1\linewidth]{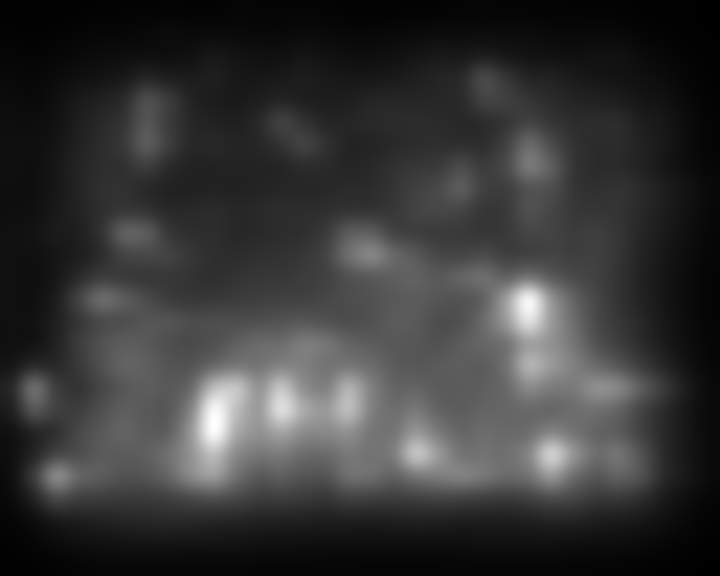} & 
  \includegraphics[width=0.1\linewidth]{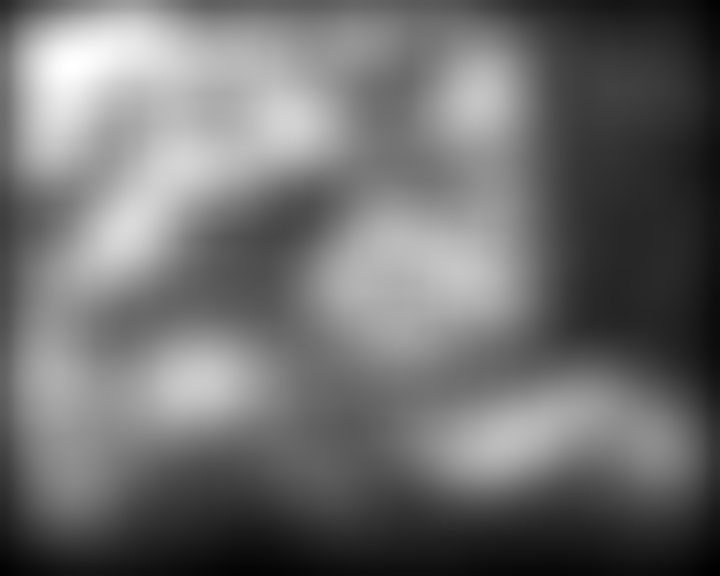} \\
  \hline
  \tiny{$\#frame 342 $ }& \includegraphics[width=0.1\linewidth]{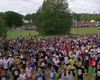} & 
  \includegraphics[width=0.1\linewidth]{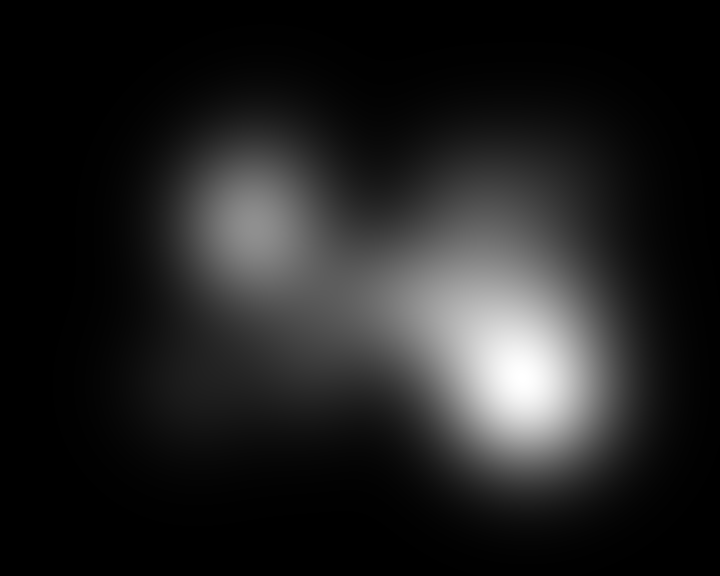} & 
  \includegraphics[width=0.1\linewidth]{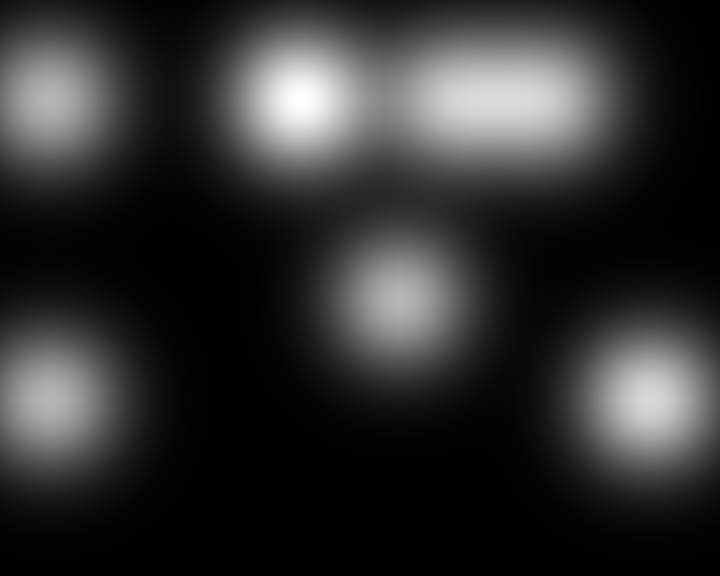} &
  \includegraphics[width=0.1\linewidth]{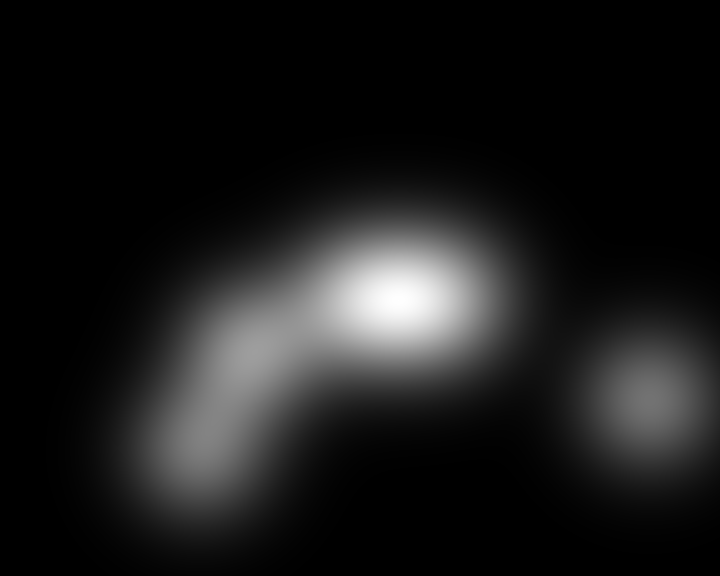} & 
  \includegraphics[width=0.1\linewidth]{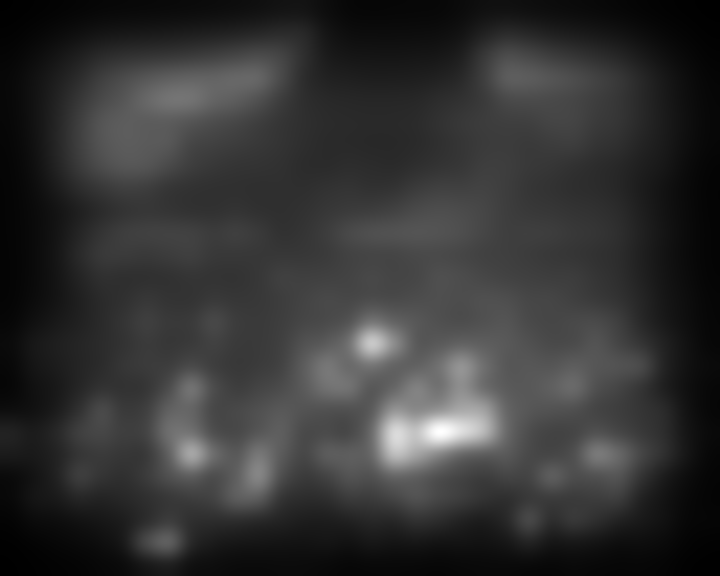} & 
  \includegraphics[width=0.1\linewidth]{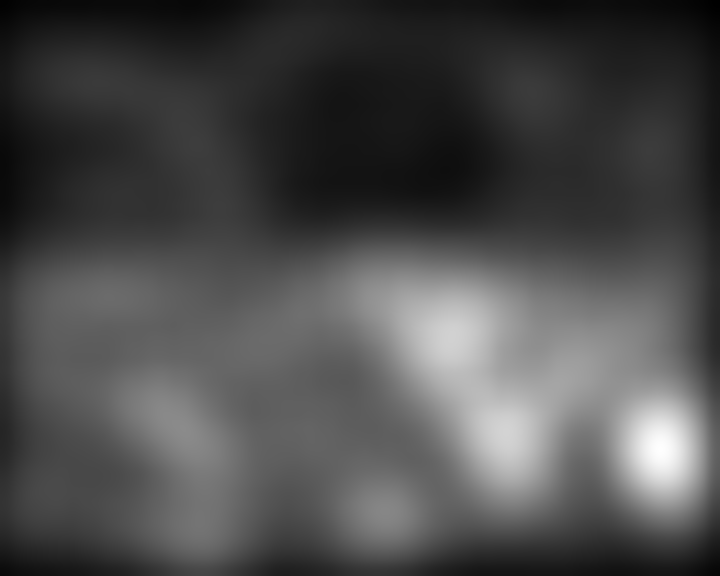} \\
  \hline
   \tiny{$\#frame 386 $} & \includegraphics[width=0.1\linewidth]{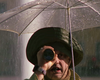} & 
  \includegraphics[width=0.1\linewidth]{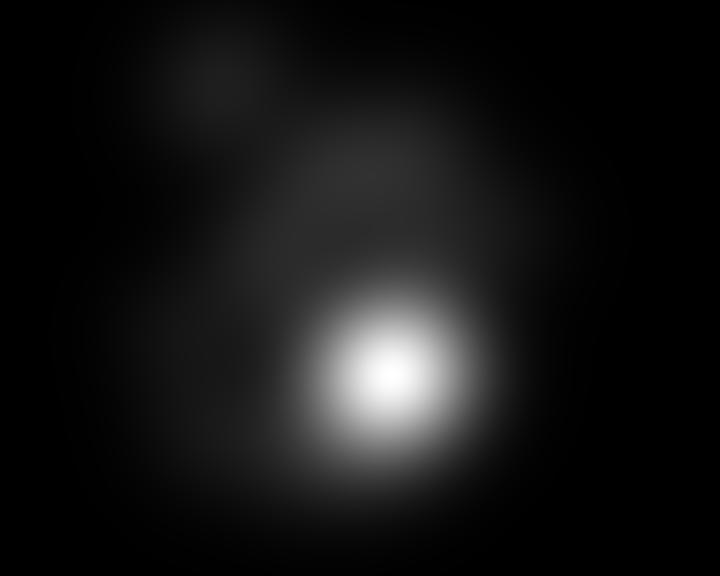} & 
  \includegraphics[width=0.1\linewidth]{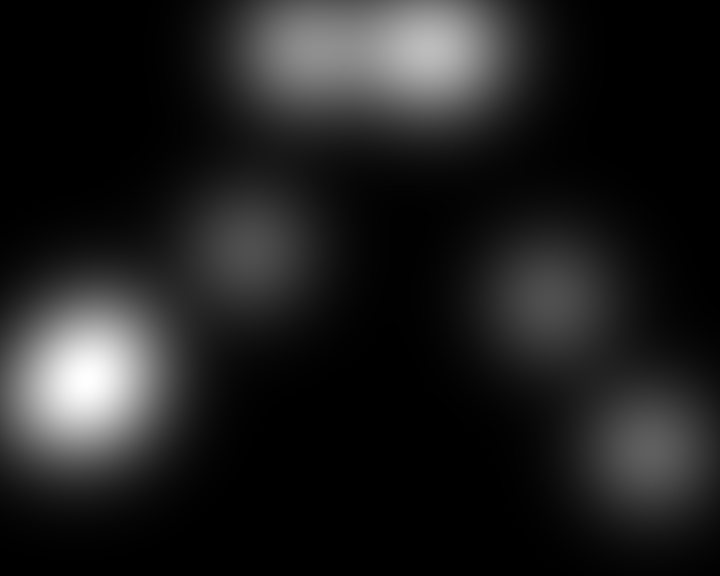} & 
  \includegraphics[width=0.1\linewidth]{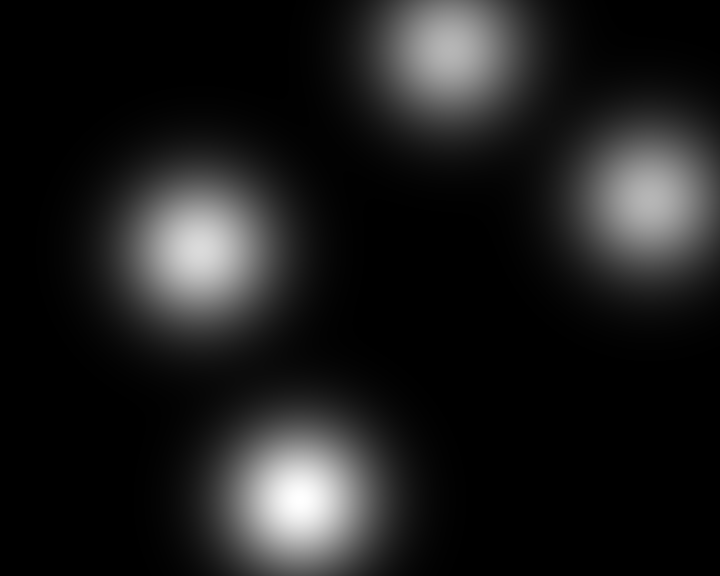} & 
  \includegraphics[width=0.1\linewidth]{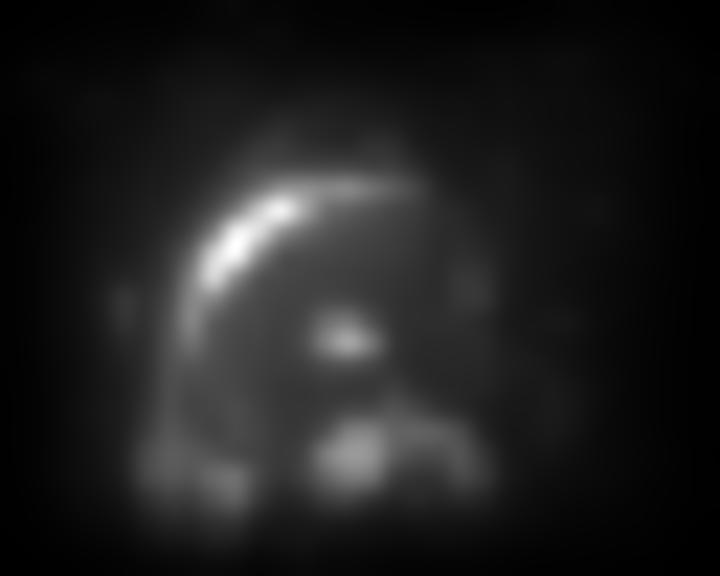} & 
  \includegraphics[width=0.1\linewidth]{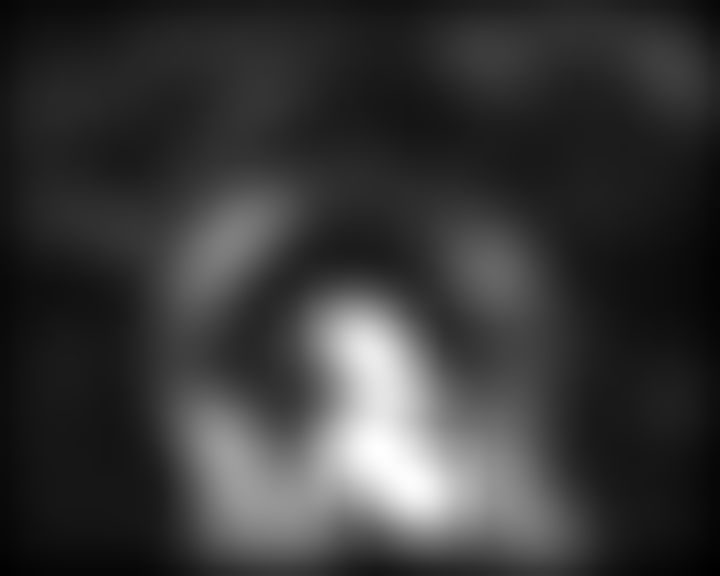} \\
  \hline
   \tiny{   $\#frame 430 $} & \includegraphics[width=0.1\linewidth]{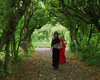} & 
  \includegraphics[width=0.1\linewidth]{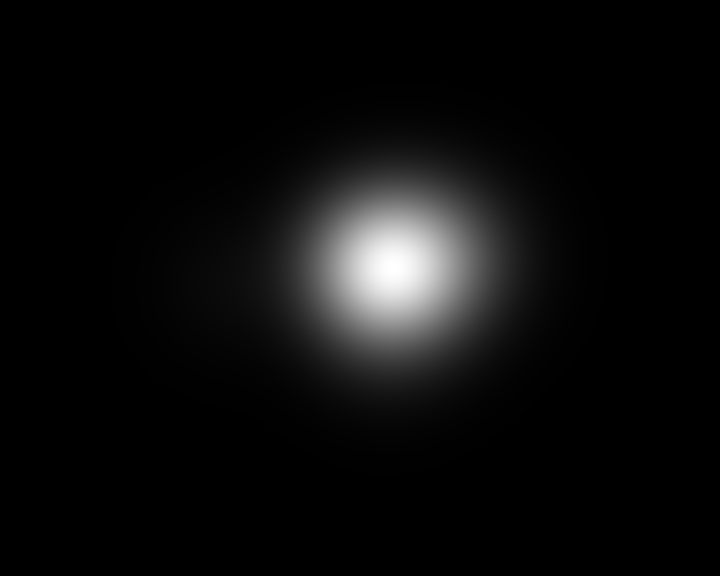} & 
  \includegraphics[width=0.1\linewidth]{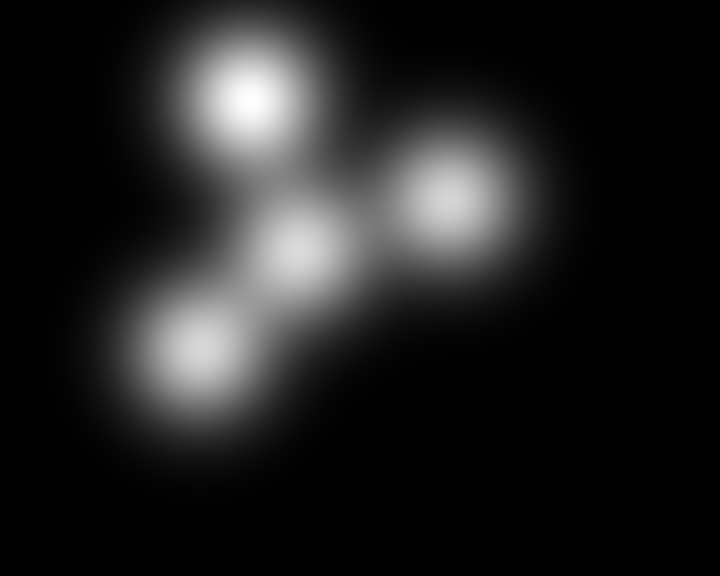} & 
  \includegraphics[width=0.1\linewidth]{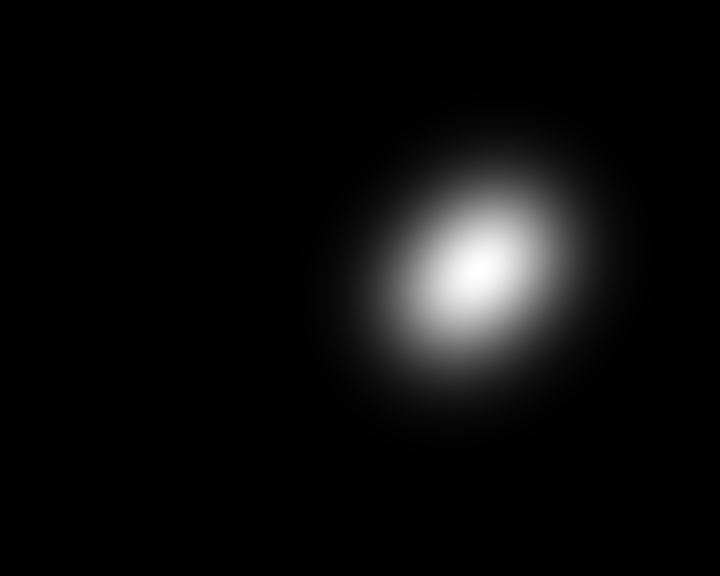} & 
  \includegraphics[width=0.1\linewidth]{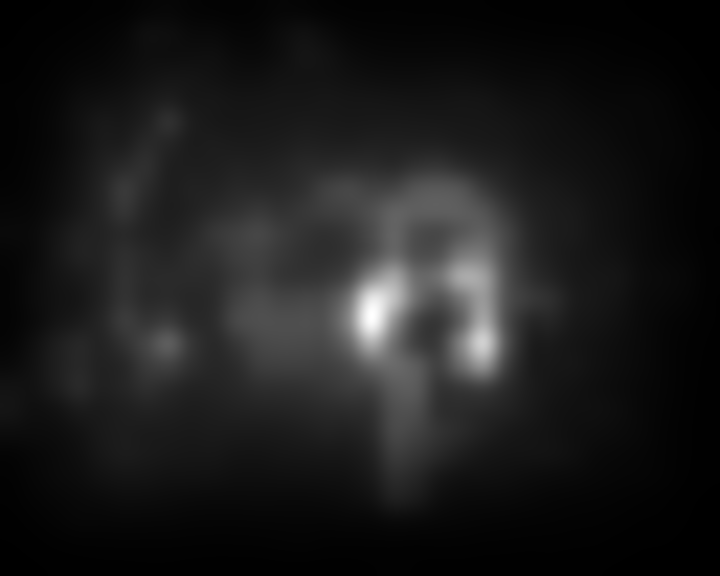} & 
  \includegraphics[width=0.1\linewidth]{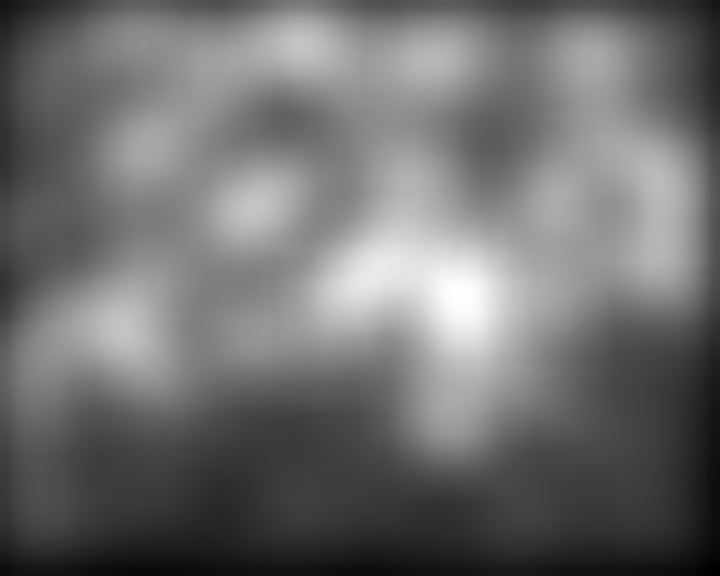} \\
  \hline
   \tiny{   $\#frame 474 $} & \includegraphics[width=0.1\linewidth]{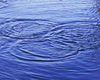} & 
  \includegraphics[width=0.1\linewidth]{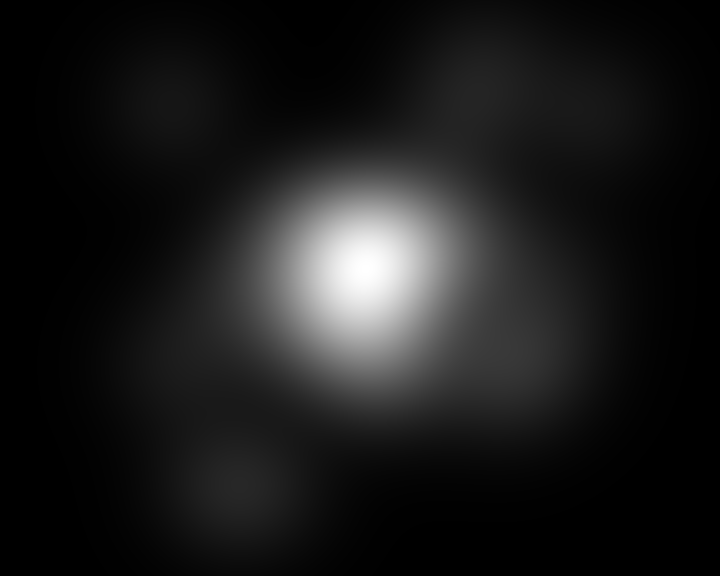} & 
  \includegraphics[width=0.1\linewidth]{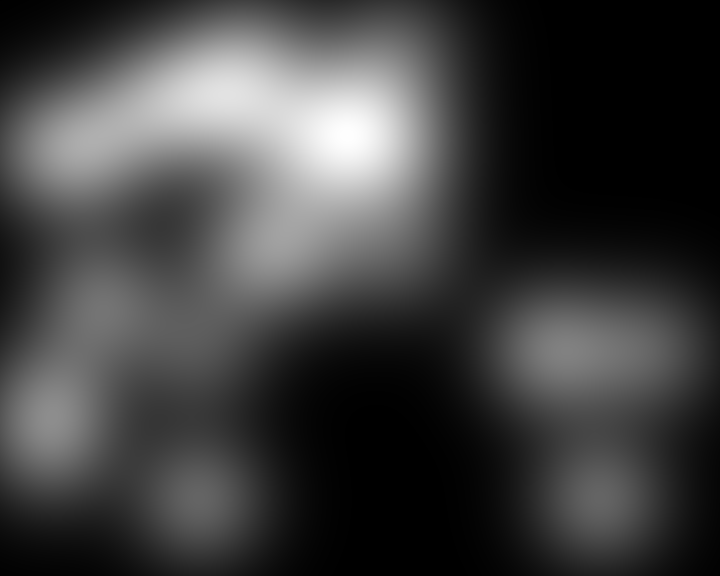} & 
  \includegraphics[width=0.1\linewidth]{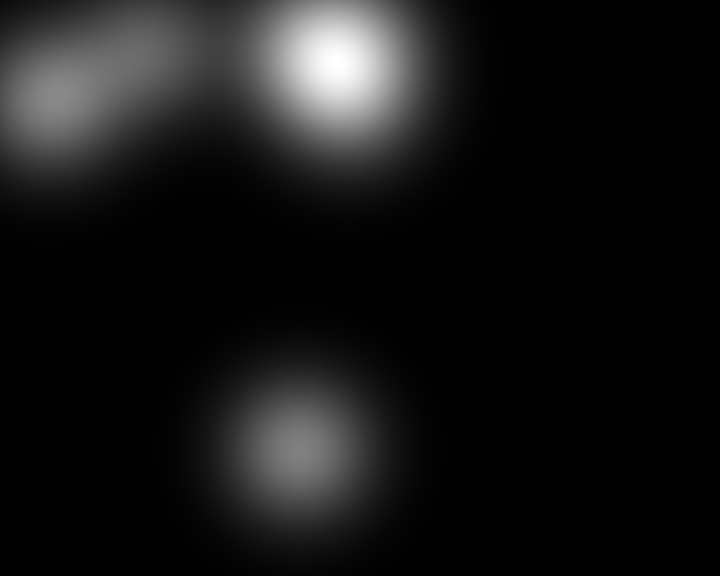} & 
  \includegraphics[width=0.1\linewidth]{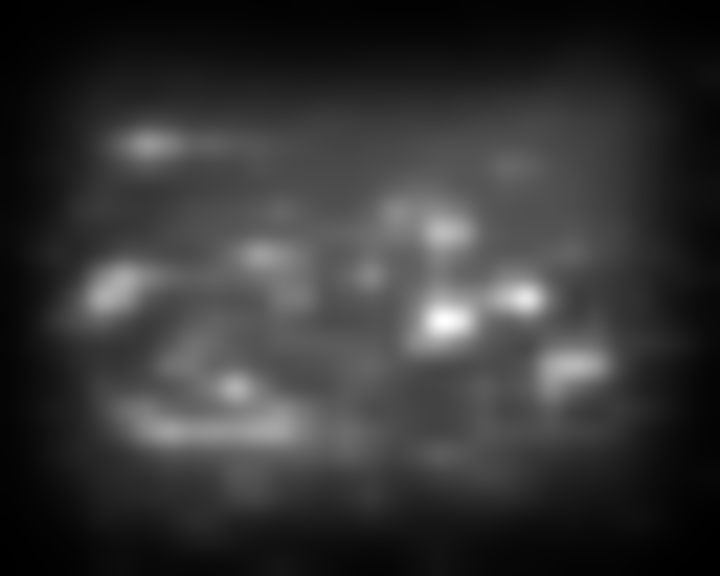} & 
  \includegraphics[width=0.1\linewidth]{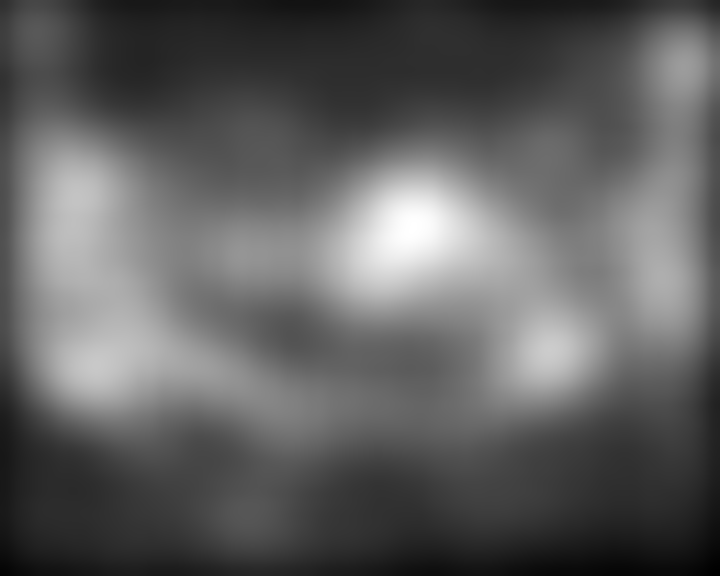} \\
  \hline
   \tiny{   $\#frame 518 $} & \includegraphics[width=0.1\linewidth]{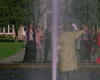} & 
  \includegraphics[width=0.1\linewidth]{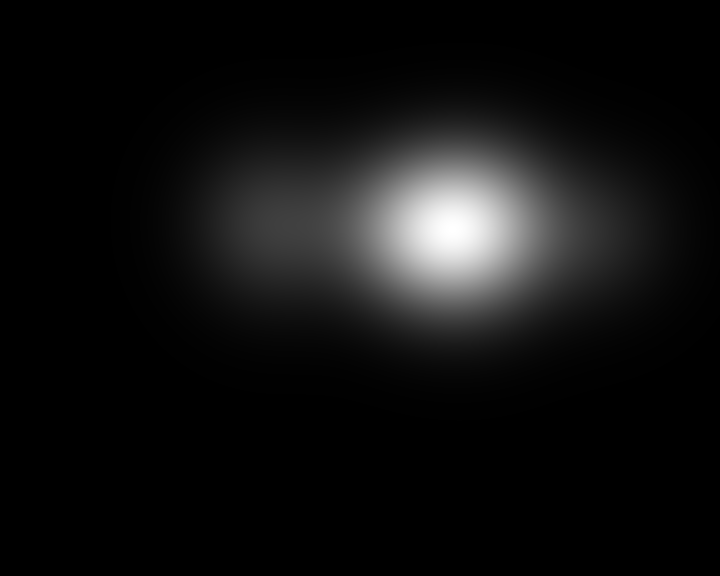} & 
  \includegraphics[width=0.1\linewidth]{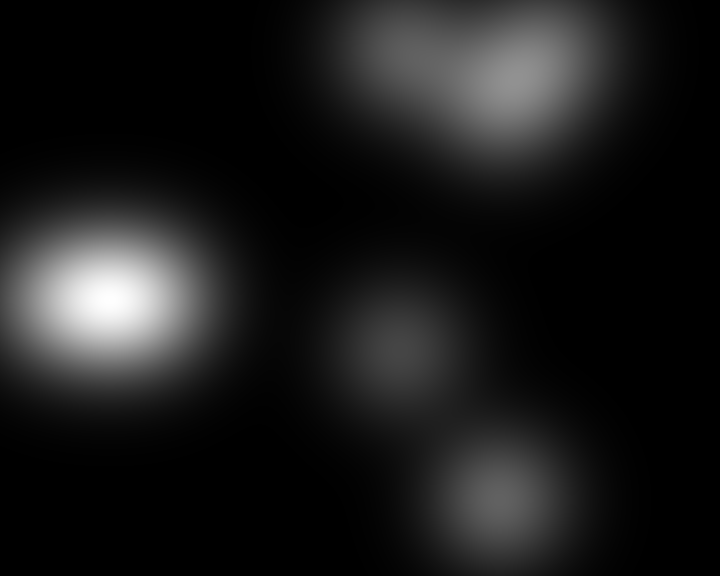} & 
  \includegraphics[width=0.1\linewidth]{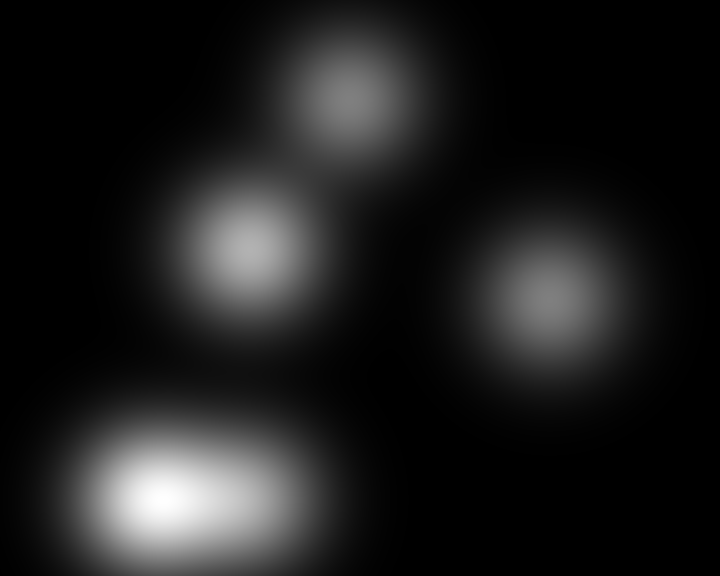} & 
  \includegraphics[width=0.1\linewidth]{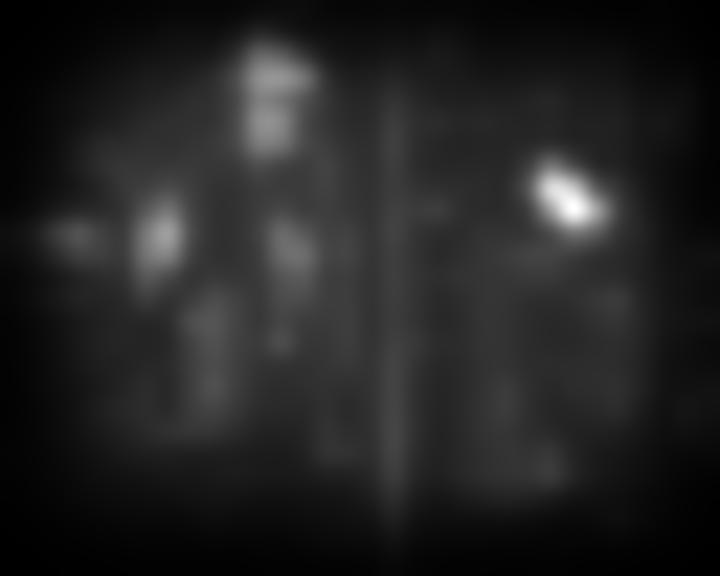} & 
  \includegraphics[width=0.1\linewidth]{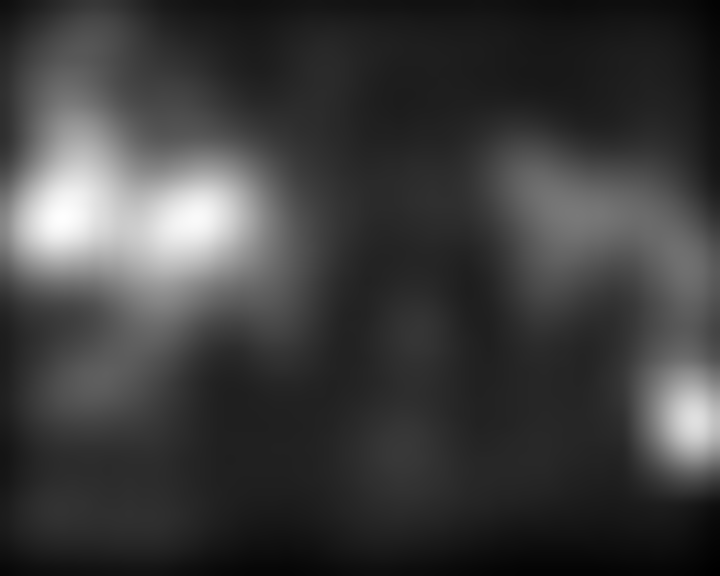} \\
  \hline
    \tiny{$\#frame 571 $} & \includegraphics[width=0.1\linewidth]{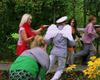} & 
  \includegraphics[width=0.1\linewidth]{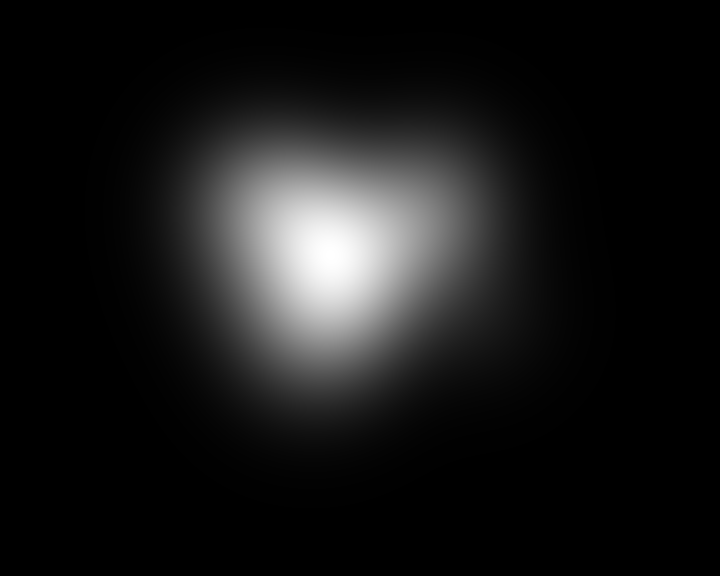} & 
  \includegraphics[width=0.1\linewidth]{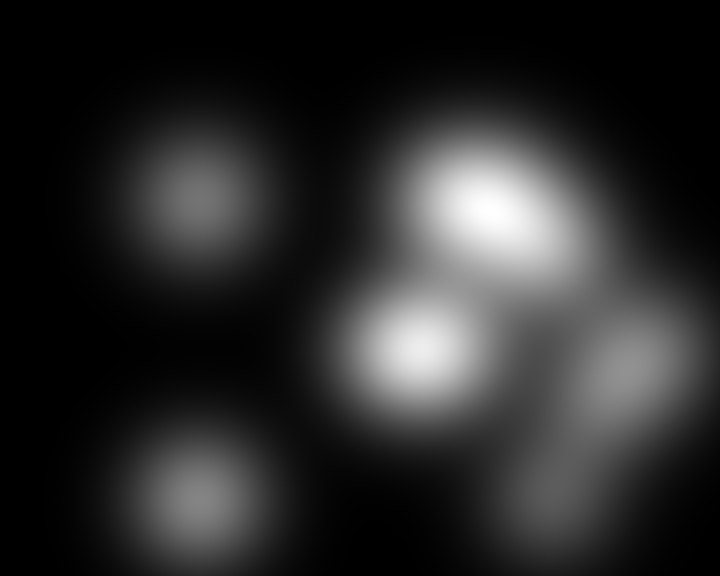} & 
  \includegraphics[width=0.1\linewidth]{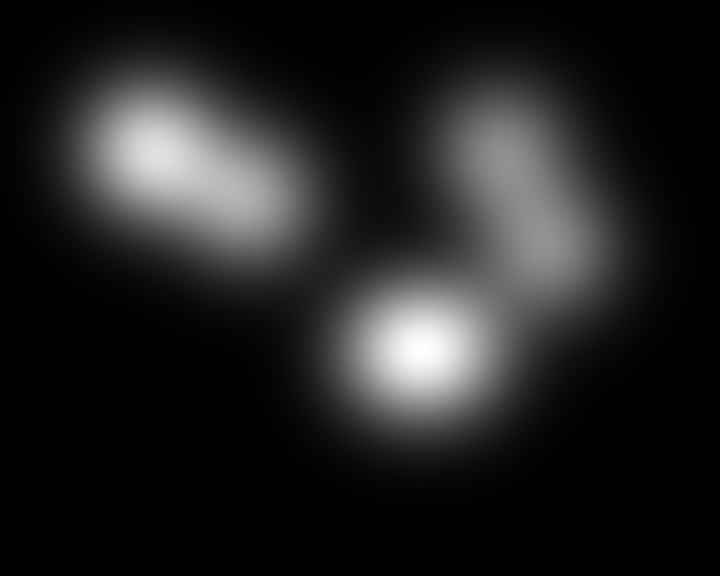} & 
  \includegraphics[width=0.1\linewidth]{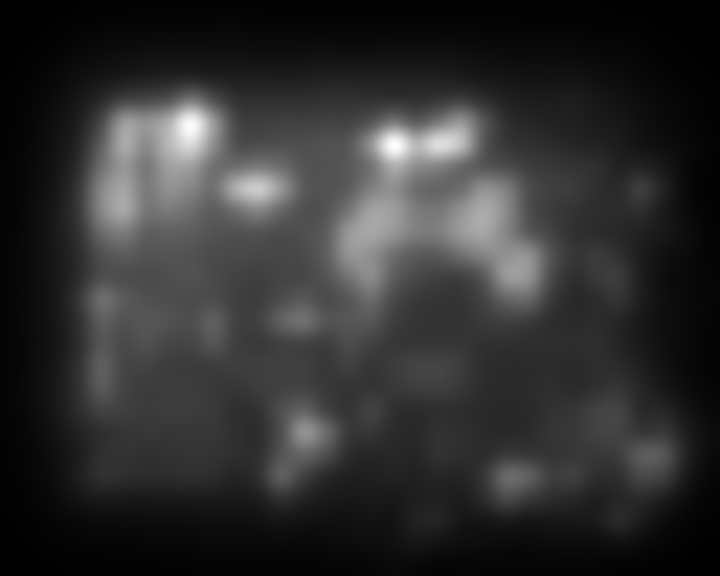} & 
  \includegraphics[width=0.1\linewidth]{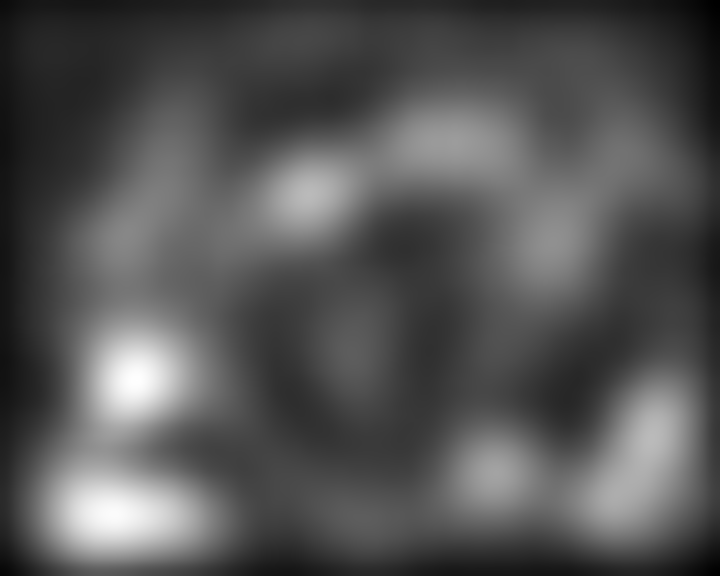} \\
  \hline
   \tiny{$\#frame 603 $} & \includegraphics[width=0.1\linewidth]{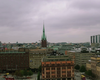} & 
  \includegraphics[width=0.1\linewidth]{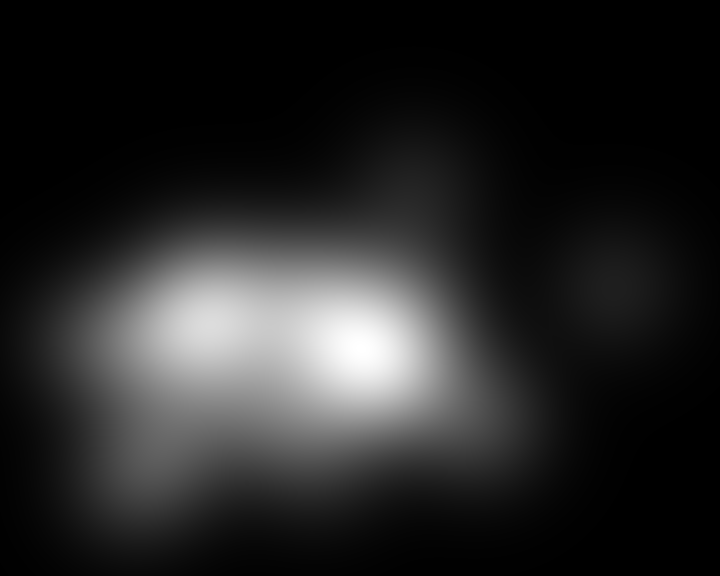} & 
  \includegraphics[width=0.1\linewidth]{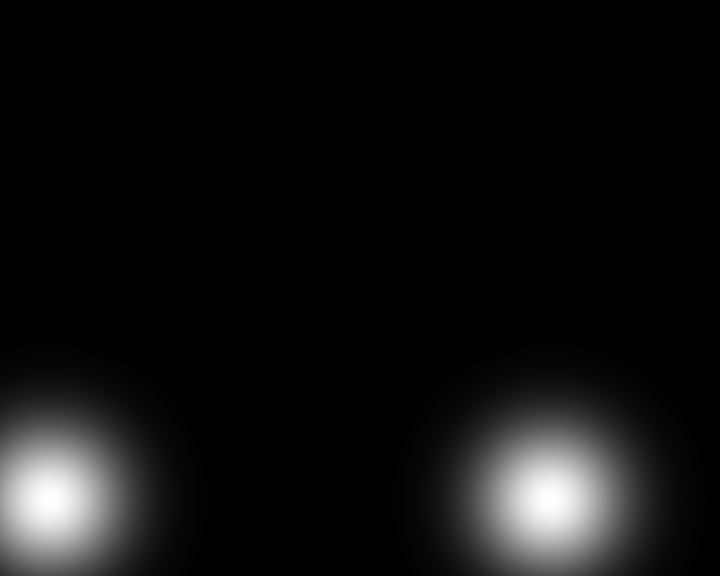} & 
  \includegraphics[width=0.1\linewidth]{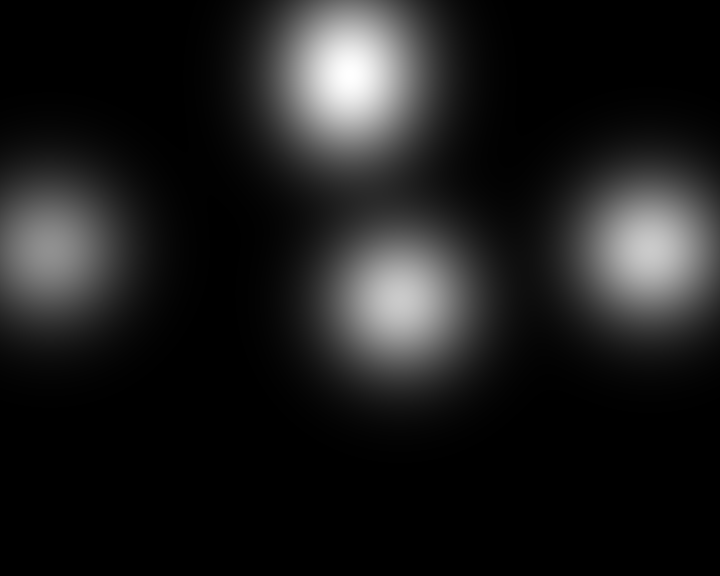} & 
  \includegraphics[width=0.1\linewidth]{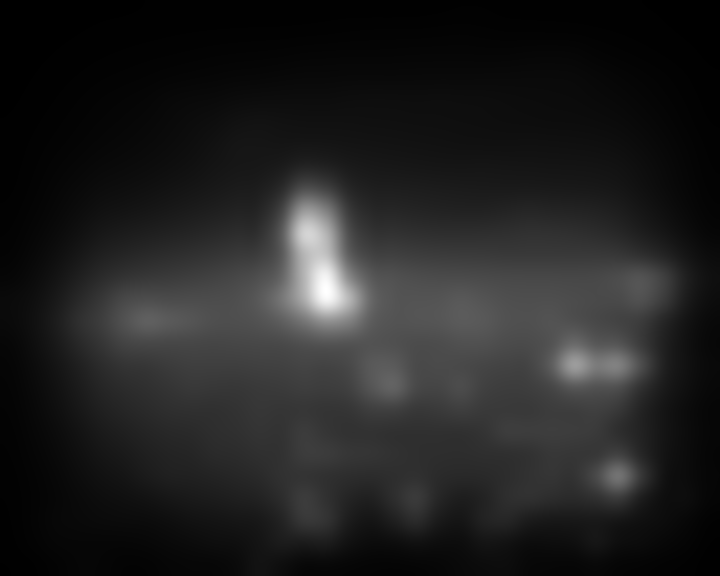} & 
  \includegraphics[width=0.1\linewidth]{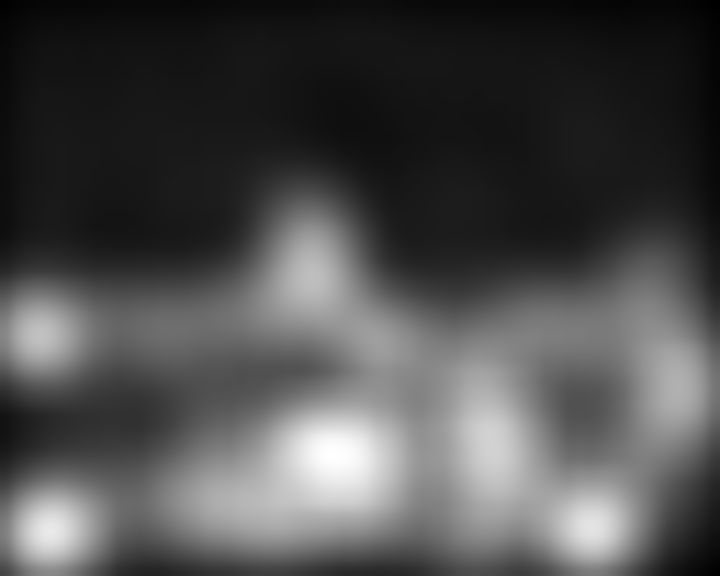} \\
  \hline
   \tiny{$\#frame 650 $} & \includegraphics[width=0.1\linewidth]{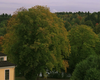} & 
  \includegraphics[width=0.1\linewidth]{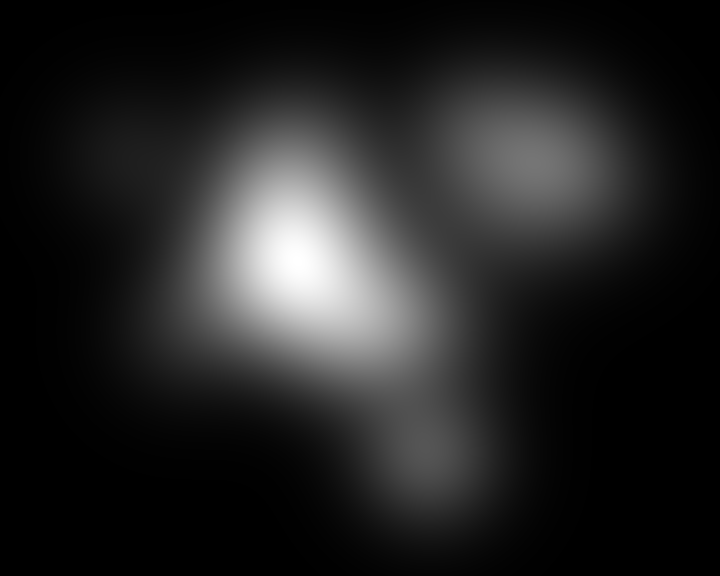} & 
  \includegraphics[width=0.1\linewidth]{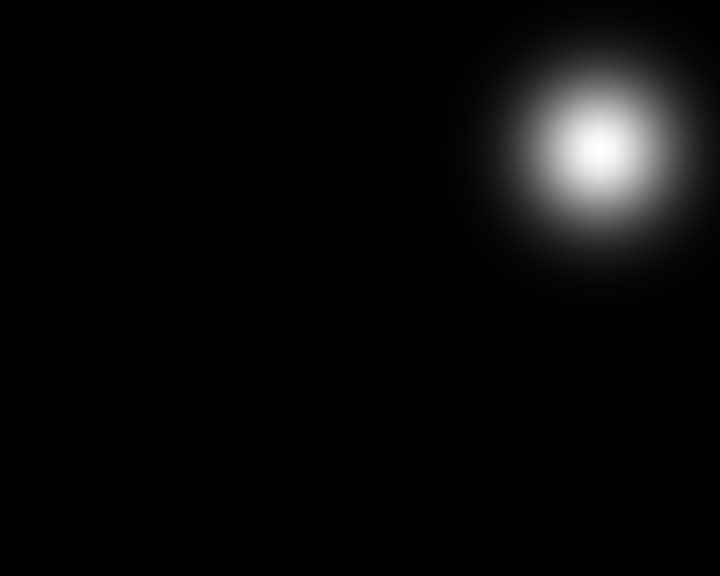} & 
  \includegraphics[width=0.1\linewidth]{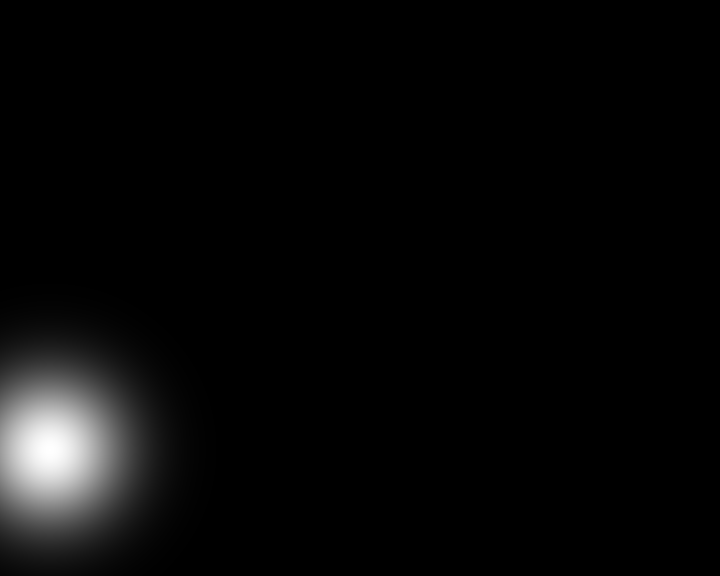} & 
  \includegraphics[width=0.1\linewidth]{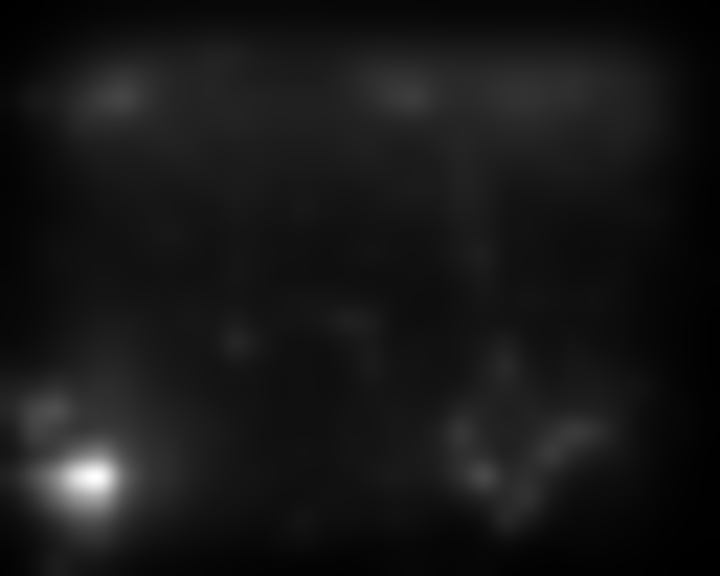} & 
  \includegraphics[width=0.1\linewidth]{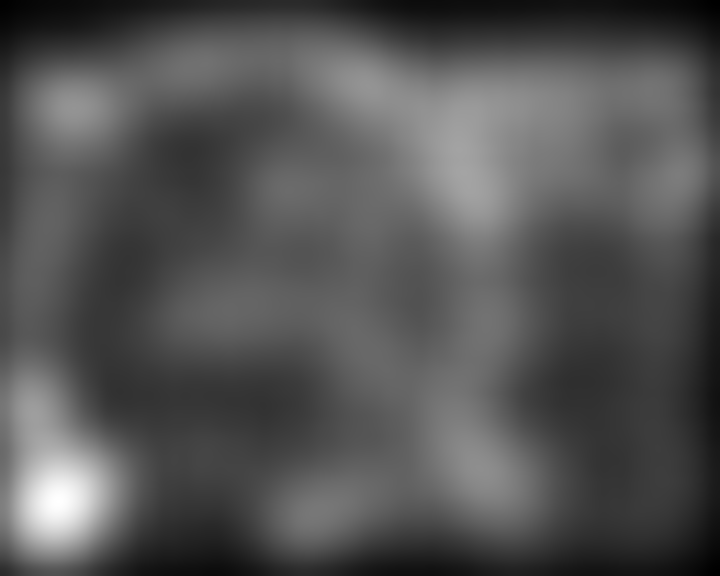} \\
  \hline
  \end{tabular}
\end{table}
\section{Experiments and results}
\label{ExperimentsAndResults}
\subsection{Datasets}
To learn the model, we have used two different datasets, the IRCCYN \cite{New43} and the HOLLYWOOD\cite{New44} \cite{New45}.

IRCCYN database contains $31$ SD videos and gaze fixations of $37$ subjects. From the overall set of $7300$ frames, 
we have extracted $6837$ salient patches and $6741$ non-salient patches.
We have used $10850$ patches ($5454$ were salient and $5396$ were non-salient) at the training step. 
For the testing step we have used $ 2728$ patches 
($1383$ salient patches and $1345$ non-salient ones) respectively.

\begin{figure}[H]
\includegraphics[width=0.8\linewidth]{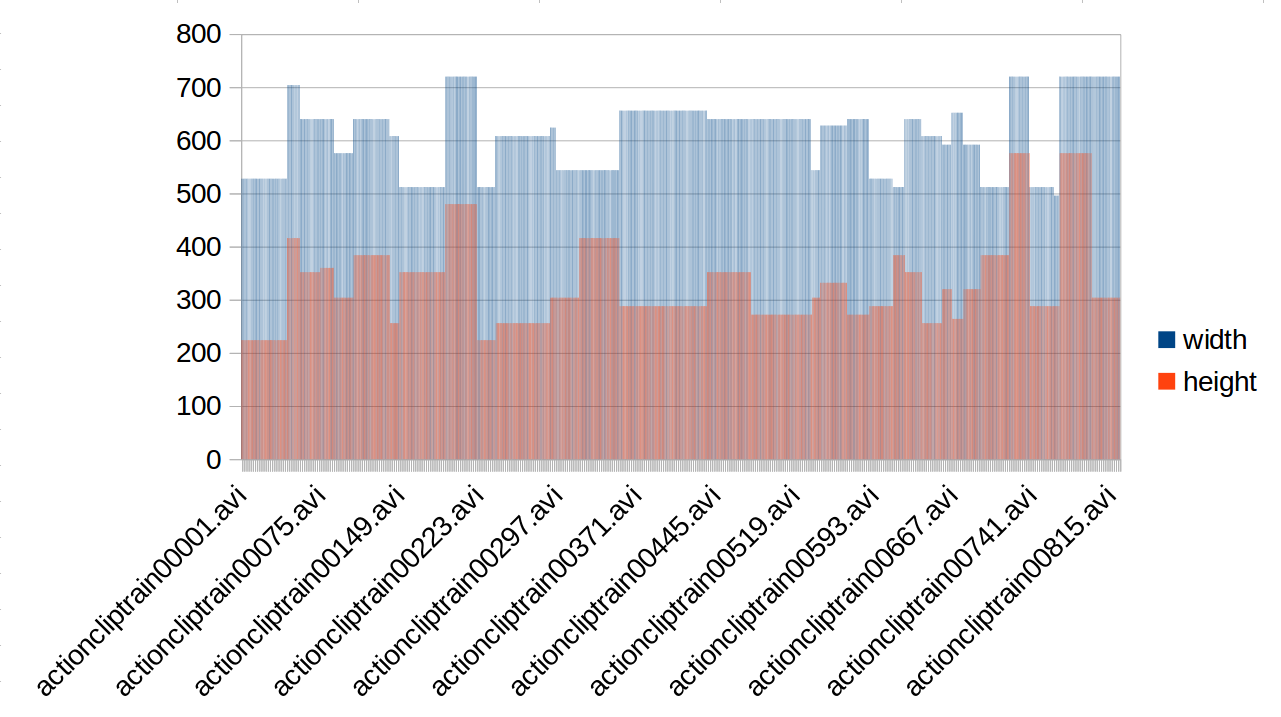}
\caption{\label{histo_hollywoodTrain} Histogram of video resolutions  $(W \times H)$ of "HOLLYWOOD" database in training step.}
\end{figure}

\begin{figure}[H]
\includegraphics[width=0.8\linewidth]{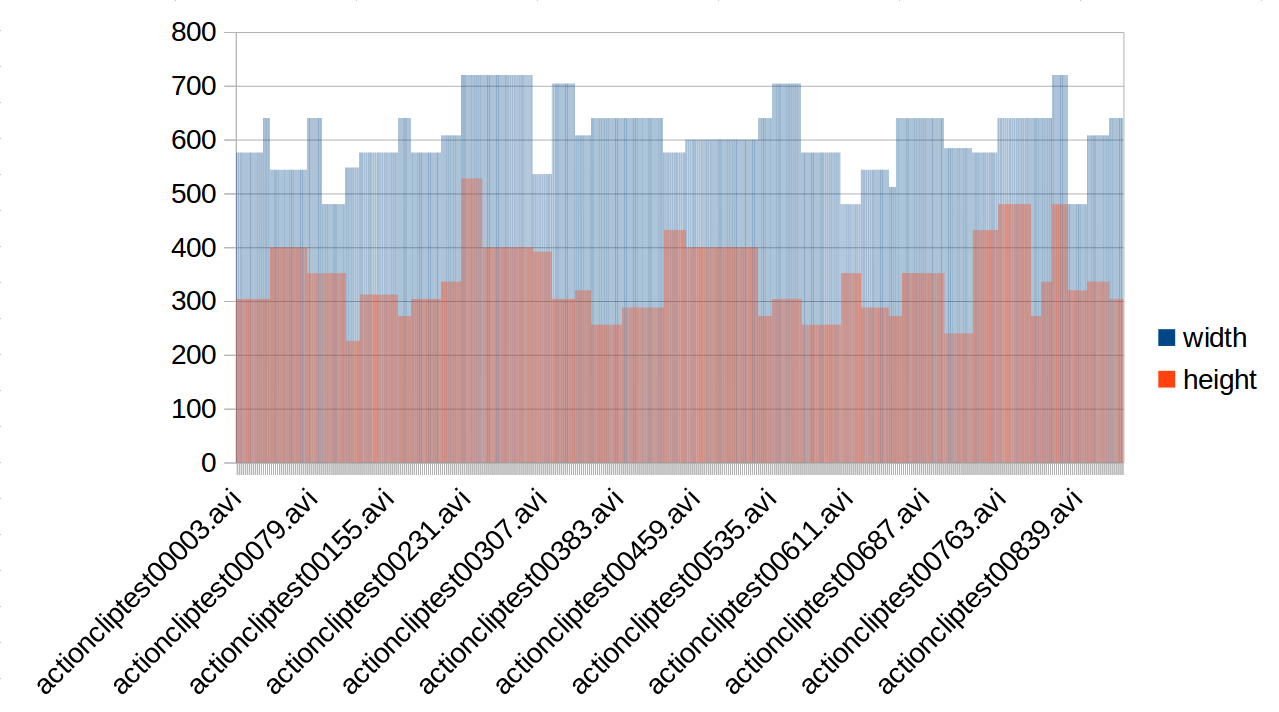}
\caption{\label{histo_hollywoodTest} Histogram of video resolutions $(W \times H)$  of "HOLLYWOOD" database in testing step.}
\end{figure}

The HOLLYWOOD database contains $823$ training videos and $884$ video for the validation step. 
The number of subjects with recorded gaze fixations varies according to each video up to $19$ subjects. 
The spatial resolution of videos varies as well. The distribution of resolutions is presented in
figures \ref{histo_hollywoodTrain} and \ref{histo_hollywoodTest}).
In another terms the HOLLYWOOD dataset contains $229825$ frames for training and  $257733$ frames for testing.
From the frames of training step we have extracted $222863$ salient patches and $221868$ non-salient patches. During the testing phase, we have used $251294$ salient patches 
and $250169$ non-salient patches respectively.

\subsection{Evaluation of patches prediction with deep CNN}

The network was implemented using a powerful graphic card Tesla K40m and processor ($2\times14$ cores). Therefore a sufficiently large amount of patches,
$256$, was used per iteration, see the $batch\_size$ parameter in equation \eqref{eq:IterationNumbers}.
After a fixed number of training iterations, a model validation step is implemented. At this stage the accuracy of the model at the current iteration is computed.

$First$ $experiment$.  To evaluate our deep network and to prove the importance of the addition of the residual motion map, we have created  two models with the same parameter settings and
architecture of the  network: the first one contained R, G and B,  primary pixel values in patches. We denote it as $DeepSaliency3k$. The $DeepSaliency4k$ presents the model using RGB 
and the normalized magnitude of residual motion as input data.
The following figures \ref{accuracy_IRCYYN} and \ref{accuracy_HOLLYWOOD} illustrate the variations of the accuracy along iterations of the both models 3k and 4k for each used
database "IRCCYN" and "HOLLYWOOD".
\begin{figure}[H]
\includegraphics[width=0.8\linewidth]{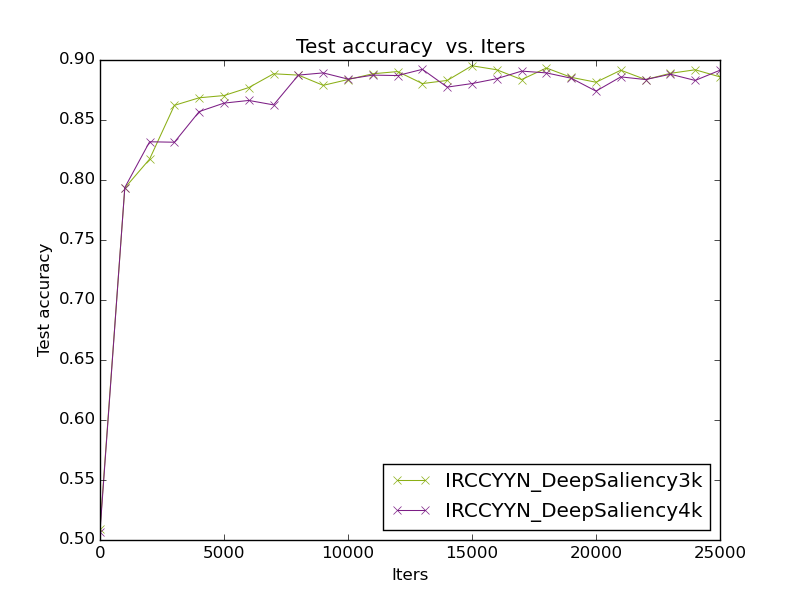}
\caption{\label{accuracy_IRCYYN} Accuracy vs iterations of the both models 3k and 4k for "IRCCYN" database.}
\end{figure}
\begin{figure}[H]
\includegraphics[width=0.8\linewidth]{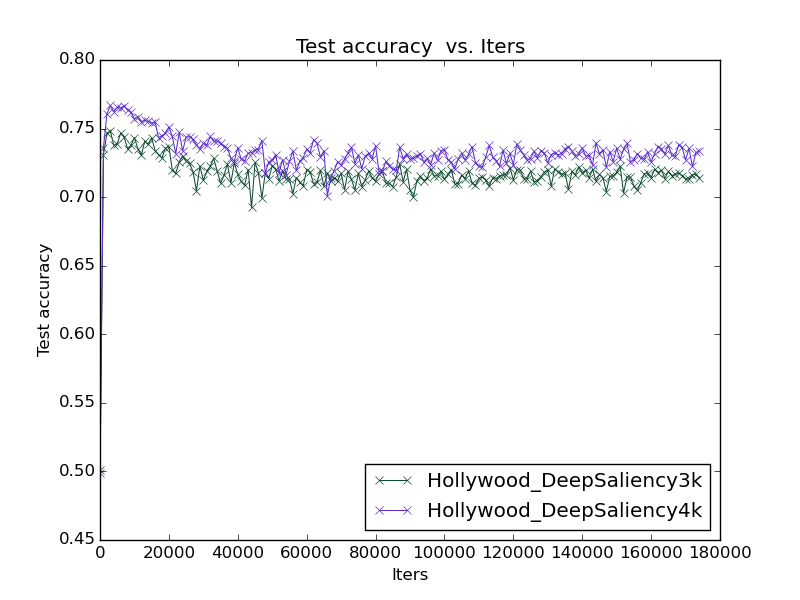}
\caption{\label{accuracy_HOLLYWOOD} Accuracy vs iterations of the both models 3k and 4k for "HOLLYWOOD" database.}
\end{figure}
\begin{table}[H]
\caption{The accuracy results on IRCCYYN and HOLLYWOOD dataset in the first experiment} 
\label{tab_accuracyResultFirstEXper}
\begin{tabular}{| c | c | c | c |}
\cline{3-4}
\multicolumn{2}{c|}{} & {3k$\_$model}& {4k$\_$model}    \\ 
\hline 
\multirow{3}{*}{ \tiny{IRCCYN} }& \tiny{$min_{(\#iter)}$} & $50.9\%_{(\#0)}$ & $50.7\%_{(\#0)}$ \\ 
\cline{2-4}
 & \tiny{$max_{(\#iter)}$ } & $89.5\%_{(\#15000)} $    & $89.2\%_{(\#13000)} $ \\ 
\cline{2-4}
 & \tiny{$avg\pm std$ } & $86.3\% \pm 0.075$ & $85.9\% \pm 0.075 $ \\ 
\hline
\multirow{3}{*}{\tiny{HOLLYWOOD}} &\tiny{$min_{(\#iter)}$}   & $50.1\%_{(\#0)}$ & $49.8\%_{(\#0)}$ \\
\cline{2-4}
 & \tiny{$max_{(\#iter)}$ } & $74.8\%_{(\#3000)}$ & $76.6\%_{(\#3000)} $\\ 
\cline{2-4}
 & \tiny{$avg\pm std$ } & $ 71.6\% \pm 0.018$  & $ 73.2\% \pm 0.020$ \\ 
\hline
\end{tabular}   
\end{table}

In the IRCCYN database, we found a higher accuracy with  both models used. The maximum value of accuracy obtained on the IRCCYN dataset is $89.5\%$ at the iteration 
$15000$ with the 3k model and $89.2\% $ at the iteration $13000$ on the 4k model, see table \ref{tab_accuracyResultFirstEXper}. We can explain 
the not improvement of the accuracy by the low number of videos in the IRCCYN dataset \cite{New50}.

For the HOLLYWOOD database, adding residual motion map improves the accuracy with almost $2\%$ on the 4k model compared to the 3k model.
The resulting accuracy of our proposed network along a fixed number 
of iterations shows the interest of adding the residual motion as a new feature together with spatial feature maps R, G and B.
Nevertheless, the essential of accuracy is obtained with purely spatial features(RGB). This is why we add spatial contrast features which have 
been proposed in classical visual saliency prediction framework \cite{New10} in the second experiment below. 

\begin{figure}[H]
\includegraphics[width=0.8\linewidth]{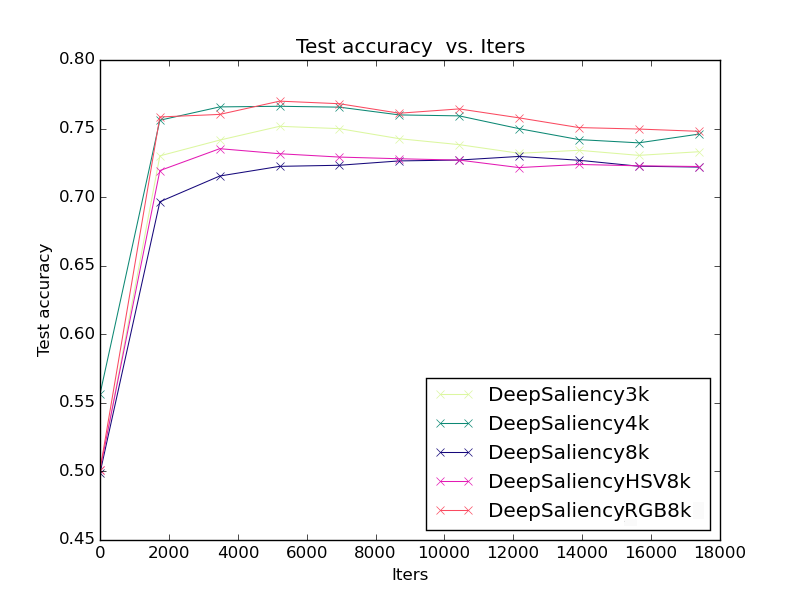}
\caption{\label{accuracyEx2_HOLLYWOOD} Second experiment: Learning of contrast feature - Accuracy vs iterations of 3k, 4k, 8k, RGB8k and HSV8k for "HOLLYWOOD" database.}
\end{figure}

$Second$ $experiment$. The second experiment  for saliency prediction is  conducted when  limiting  the maximal number of iterations to prevent us from falling into overfitting problem. 
Instead of increasing the number of training iterations with a limited number of data samples before each validation iteration, as this is the case in the work of \cite{New33}, 
we pass all the training set before the validation of the parameters and limit the maximal number of iterations in the whole training process. 
This drastically decreases (12 times approximately) the training complexity, without the loss of accuracy (see tables \ref{tab_accuracyResultFirstEXper} and \ref{tab_accuracyResultSecondEXper} for 3k and 4k models). 
In order to evaluate the performance of contrast features in a deep learning spatio-temporal model, we test the "8K" model first. Its input layers are composed of $7$ contrast features,
as described in section \ref{PrimarySpatialFeatures} and of the residual motion map. The results are presented in table \ref{tab_accuracyResultSecondEXper}  and illustrated in figure
\ref{accuracyEx2_HOLLYWOOD}. 
It can be seen, that contrasts features only combined with motion yield poorer performance with regard to 3K and 4K models.
Therefore, we keep primary colour information in the further 
HSV8K and RGB8K models.

\begin{table}[H]
\caption{The accuracy results on HOLLYWOOD dataset during the second experiment}
\label{tab_accuracyResultSecondEXper}
\begin{tabular}{| c | c | c | c | c | c |}
\cline{2-6}
 \multicolumn{1}{c|}{} & \tiny{3k$\_$model}& \tiny{4k$\_$model} & \tiny{8k$\_$model} & \tiny{RGB8k$\_$model} & \tiny{HSV8k$\_$model}     \\
\hline
\tiny{$min_{(\#iter)}$} & \tiny{$49.8\%_{(\#0)}$ }& \tiny{$55.6\%_{(\#0)}$ }& \tiny{$49.8\%_{(\#0)}$} & \tiny{$50.1 \%_{(\#0)}$ }& \tiny{$ 50.1\%_{(\#0)}$}\\
\hline
\tiny{$max_{(\#iter)}$ } & \tiny{$75.1\%_{(\#5214)}$ }& \tiny{$76.6\%_{(\#5214)}$ }& \tiny{$ 72.9\%_{(\#12166) }$}& \tiny{$76.9 \%_{(\#5214 )}$}& \tiny{$ 73.5\%_{(\#3476) }$}\\
\hline
\tiny{$avg\pm std$ } & \tiny{$ 71.6\% \pm 0.072$}& \tiny{$ 73.6\% \pm 0.060$} &\tiny{ $ 70.1\% \pm 0.067 $}& \tiny{$ 73.5\% \pm 0.078$ }& \tiny{$ 70.5\% \pm 0.068 $}\\
\hline
\end{tabular}  
\end{table}

\subsection{Evaluation of predicted visual saliency maps}

In the literature, various evaluation criteria were used to determine the level of similarity between visual attention maps and gaze fixations of subjects like the
normalized scanpath 
saliency 'NSS', Pearson Correlation Coefficient 'PCC', and the area under the ROC curve  'AUC'  \cite{New20}\cite{New21}.
The «Area under the ROC Curve» measures the precision and accuracy of a system with the goal of categorizing entities 
into two distinct groups based on their features. The image pixels may belong either to the category of pixels fixated by subjects,
either to the category of pixels that are not fixated  by any subject. More the area is large, more the curve deviates from the line of the random
classifier (area $ 0.5 $) and approaches to the ideal bend of the classifier (area $ 1.00 $).
A value of AUC close to $ 1 $ indicates a correspondence between the predicted saliency and the eye positions. While a value close to $ 0.5 $ presents
a random generation of the salient areas by the model computing the saliency maps. Therefore the objective and subjective saliency 
differs strongly. 
In our work, visual saliency being predicted by a deep CNN classifier, we have computed the hybrid AUC metric between predicted saliency maps
and gaze-fixations as in \cite{New53}. 
The results of the experiments are presented in the tables 6 and 7 below on an arbitrary chosen subset of 12 videos from HOLLYWOOD dataset.
The figures depicted in the tables 
correspond to the maximum value obtained during the training and validation (as presented in tables 4 and 5). For the first experiment the maximal number
of iterations was set to 
174000 and for the second experiment, this number was fixed $10$ times lower. From table 6 it can be stated that i)adding primary motion features, such as residual motion improves the 
quality of predicted visual attention maps whatever is the training of the network. The improvement is systematic and goes up to $38\%$  in case of 
clipTest105 
(in the first experiment);ii) the way to train the network, we propose with lower number of iterations and all training data used does not strongly affect the performances. 
Indeed, with 4k model the results are better for almost all clips, see highlighted figures in table 6.  
In table 7 we compare all our predicted saliency models with gaze fixations. It comes out that more complex models yield better results:
up to $42\%$ of improvement in clipTest250. The quality of the prediction of patches (see table 4, 5 and figure 11) DeepRGB8K outperforms DeepHSV8k.
Therefore, for comparison with reference models from the state of the art, $GBVS$, $SignatureSal$ and spatio-temporal model 
by Seo \cite{New31}, named "Seo" we use $DeepRGB8K$ model, 
see table \ref{comparaison_withReference} below.

\begin{table}[H]
\caption{The comparison, with AUC metric, of the two experiments for 3K and 4K saliency models vs gaze fixations 'Gaze-fix' on a subset of HOLLYWOOD dataset} 
\label{comparaison_twoexperiment}
\begin{tabular}{ | c | c | c | c | c | }
\cline{2-5}
\multicolumn{1}{c|}{} & \multicolumn{2}{c|}{ First Experiment} & \multicolumn{2}{c|}{ Second Experiment} \\ 
\hline
\tiny{ VideoName } & \tiny{ Gaze-fix vs Deep3k} & \tiny{ Gaze-fix vs Deep4k} & \tiny{ Gaze-fix vs Deep3k} & \tiny{ Gaze-fix vs Deep4k} \\ 
\hline 
\tiny{clipTest1 }&\tiny{ $0,58612\pm 0,19784$} &\tiny{ $0,61449\pm0,17079$ } & \tiny{ $ 0,55641\pm0,20651 $} & \tiny{\underline{$ 0,77445\pm0,14233$}}  \\ 
\hline 
\tiny{clipTest56 }&\tiny{ $0,74165\pm0,17394$} & \tiny{ $0,75911\pm0,12509 $ } & \tiny{ $0,65480\pm0,19994 $ }&\tiny{\underline{  $ 0,82034\pm0,12727$}} \\
\hline 
\tiny{clipTest105 }&\tiny{ $0,35626\pm0,33049$} & \tiny{ $0,74312\pm0,19479 $ } & \tiny{ $0,66285\pm0,20553 $ }&\tiny{ \underline{$ 0,74740\pm0,14689$}} \\
\hline 
\tiny{ClipTest200}&\tiny{ $0,50643\pm0,241466$ }& \tiny{ $0,59407\pm0,20188 $} & \tiny{  $0,53926\pm0,21976 $} & \tiny{ \underline{$0,69309\pm0,16428 $}} \\
\hline
\tiny{ClipTest250}&\tiny{ $0,548647\pm0,240311$ }& \tiny{\underline{ $0,754679\pm0,15476 $}}& \tiny{  $0,41965\pm0,28409 $} &\tiny{  $0,72621\pm0,15028 $} \\
\hline
\tiny{ClipTest300}&\tiny{ $0,58236\pm0,22632$ }& \tiny{ $0,66156\pm0,16352 $} & \tiny{  $0,33808\pm0,19672 $} &\tiny{ \underline{ $0,79186\pm0,09732 $}}  \\
\hline
\tiny{ClipTest350}&\tiny{ $0,67679\pm0,29777$ }& \tiny{ $0,739803\pm0,16859 $} & \tiny{  $0,47971\pm0,40607 $} &\tiny{ \underline{ $0,80467\pm0,15750 $}}\\
\hline
\tiny{ClipTest500} & \tiny{ $ 0,58351\pm0,20639 $} &\tiny{  $0,75242\pm0,15365 $  } & \tiny { $0,36761\pm0,36777 $} &\tiny{ \underline{ $0,82230\pm0,15196 $}} \\
\hline
\tiny{ClipTest704 }& \tiny{ $0,59292\pm0,18421 $} &\tiny{  $0,68858\pm0,16278 $  } & \tiny {  $0,46192\pm0,21286$} &\tiny{  \underline{$ 0,76831\pm0,11186$}} \\
\hline
\tiny{ClipTest752} & \tiny{ $0,41710\pm0,11422 $} &\tiny{ \underline{ $0,63240\pm0,16870 $ } } & \tiny {  $0,25651\pm0,25830$} &\tiny{  $0,58621\pm0,21568 $} \\
\hline
\tiny{ClipTest803} & \tiny{ $0,67961\pm0,24997 $} &\tiny{  $0,82489\pm0,14023 $  } & \tiny {  $0,55019\pm0,18646 $} &\tiny{  \underline{$0,87474\pm0,06946 $}} \\
\hline
\tiny{ClipTest849} & \tiny{ $ 0,39952\pm0,31980$} &\tiny{  $0,67103\pm0,20623 $  } & \tiny {  $0,30190\pm0,27491 $} &\tiny{  \underline{$0,81148\pm0,10363 $}} \\
\hline
\end{tabular}   
\end{table}

\begin{table}[H]
\caption{The comparison metric of gaze fixations 'Gaze-fix' vs Deep saliency '3k', '4k', '8k' , 'RGB8k' and 'HSV8k') for the video from HOLLYWOOD } 
\label{tab3}
\begin{tabular}{ | c | c | c | c | c | c | }
\hline
\tiny{ VideoName } & \tiny{Gaze-fix vs Deep3k} & \tiny{Gaze-fix vs Deep4k} & \tiny{Gaze-fix vs Deep8k} & \tiny{Gaze-fix vs DeepRGB8k} & \tiny{Gaze-fix vs DeepHSV8k} \\ 
\hline 
\tiny{clipTest1} & \tiny{ $ 0,55641\pm0,20651 $} & \tiny{\underline{ $ 0,77445\pm0,14233$}} & \tiny{ $0,58518\pm0,17991 $} & \tiny{ $0,725073\pm0,168168 $} &\tiny{ $0,76923\pm0,09848 $ } \\ 
\hline 
\tiny{clipTest56} & \tiny{ $0,65480\pm0,19994 $ }&\tiny{  $ 0,82034\pm0,12727$} &\tiny{  $0,78106\pm0,090992 $ }& \tiny{\underline{ $0,82244\pm0,07295 $}} & \tiny{ $ 0,81651\pm0,06100 $ }\\
\hline 
\tiny{ClipTest105} & \tiny{ $0,66285\pm0,20553 $ }&\tiny{ $ 0,74740\pm0,14689$}&\tiny{  $0,71597\pm0,11538 $ }&\tiny{  $0,63652\pm0,22207 $} & \tiny{\underline{ $0,81365\pm0,08808  $}}\\
\hline 
\tiny{ClipTest200}&\tiny{  $0,53926\pm0,21976 $} & \tiny{ $0,69309\pm0,16428 $} & \tiny{ $0,74225\pm0,19740 $ }& \tiny{\underline{ $0,77948\pm0,17523 $ }}& \tiny{ $0,68396\pm0,17425 $} \\
\hline
\tiny{ClipTest250}&\tiny{  $0,41965\pm0,28409 $} &\tiny{  $0,72621\pm0,15028 $} & \tiny{ $0,51697\pm0,21393 $ }& \tiny{\underline {$0,84299\pm0,10787 $ }}& \tiny{ $0,69886\pm0,13633 $} \\
\hline
\tiny{ClipTest300}&\tiny{  $0,33808\pm0,19672 $} &\tiny{  $0,79186\pm0,09732 $} & \tiny{ $0,79265\pm0,10030 $ }& \tiny{ $0,74878\pm0,12161 $ }& \tiny{\underline{ $0,83009\pm0,08418 $}} \\
\hline
\tiny{ClipTest350}&\tiny{  $0,47971\pm0,40607 $} &\tiny{ \underline{ $0,80467\pm0,15750 $}} & \tiny{ $0,78924\pm0,16506 $ }& \tiny{ $0,72284\pm0,16996 $ }& \tiny{ $0,80009\pm0,232312$} \\
\hline
\tiny{ClipTest500}&\tiny{  $0,36761\pm0,36777 $} &\tiny{  $0,82230\pm0,15196 $} & \tiny{ $0,68157\pm0,15676 $ }& \tiny{ $0,85621\pm0,16137 $ }& \tiny{\underline{ $0,88067\pm0,09641 $}} \\
\hline
\tiny{ClipTest704}&\tiny {  $0,46192\pm0,21286$} &\tiny{  $ 0,76831\pm0,11186$}& \tiny{\underline{ $0,80725\pm0,11455$ }}& \tiny{ $0,78256\pm0,09523 $ }& \tiny{\underline{ $0,79551\pm0,071867 $}} \\
\hline
\tiny{ClipTest752}&\tiny {  $0,25651\pm0,25830$} &\tiny{  $0,58621\pm0,21568 $} & \tiny{ \underline{$0,78029\pm0,08851 $ }}& \tiny{ $0,59356\pm0,17804$ }& \tiny{  $0,76665\pm0,07837   $} \\
\hline
\tiny{ClipTest803}&\tiny {  $0,55019\pm0,18646 $} &\tiny{  $0,87474\pm0,06946 $} & \tiny{ $0,84338\pm0,06868 $ }& \tiny{\underline{ $0,88170\pm0,10827 $} }& \tiny{ $0,85641\pm0,06181 $} \\
\hline
\tiny{ClipTest849}&\tiny {  $0,30190\pm0,27491 $} &\tiny{  $0,81148\pm0,10363 $} & \tiny{ $0,70777\pm0,08441 $ }& \tiny{ \underline{$0,91089\pm0,05217 $} }& \tiny{ $0,71224\pm0,07434 $} \\
\hline
\end{tabular}   
\end{table}
 
\begin{table}[H]
\caption{The comparison of AUC metric gaze fixations 'Gaze-fix' vs predicted saliency 'GBVS', 'SignatureSal' and 'Seo') and our DeepRGB8k for the
videos from HOLLYWOOD dataset} 
\label{comparaison_withReference}
\begin{tabular}{ | c | c | c | c | c | }
\hline
\tiny{VideoName} & \tiny{Gaze-fix vs GBVS} & \tiny{Gaze-fix vs SignatureSal} & \tiny{Gaze-fix vs Seo} & \tiny{Gaze-fix vs DeepRGB8k}\\ 
\hline 
\tiny{clipTest1}  & \underline{\tiny{ $ 0,81627 \pm0,10087 $}} & \tiny{ $0,69327 \pm 0,13647$} & \tiny{ $0,50090 \pm 0,06489 $} & \tiny{$0,725073\pm0,168168 $ }\\ 
\hline 
\tiny{clipTest56 } & \tiny{ $ 0,76594\pm0,11569$ }&\tiny{  $0,75797\pm0,08650$} &\tiny{  $0,64172\pm0,11630 $ } & \tiny{\underline{$0,82244\pm0,07295 $} }\\
\hline 
\tiny{clipTest105 } & \tiny{$ 0,63138\pm0,16925$ }&\tiny{  $0,57462\pm0,13967$} &\tiny{  $0,54629\pm0,12330$ } & \tiny{\underline{$0,63652\pm0,22207 $ }}\\
\hline
\tiny{clipTest200 } &\tiny{  $0,75904\pm0,17022$} &\tiny{ \underline{ $0,87614\pm0,10807$}} & \tiny{ $0,65675\pm0,13202$ } & \tiny{$0,77948\pm0,17523 $ }\\
\hline
\tiny{clipTest250 } &\tiny{  $0,74555\pm0,09992$} &\tiny{  $0,69339\pm0,11066$} & \tiny{ $0,47032\pm0,10193$ } & \tiny{\underline{$0,84299\pm0,10787 $ }} \\
\hline
\tiny{clipTest300} &\tiny{\underline{$0,82822\pm0,11143$}} &\tiny{  $0,81271\pm0,12922$} & \tiny{ $0,75965\pm0,13658$ } & \tiny{$0,74878\pm0,12161 $ } \\
\hline
\tiny{clipTest350} &\tiny{  $0,65136\pm0,16637$} &\tiny{ $0,68849\pm0,249027$} & \tiny{ $0,57134\pm0,12408$ } & \tiny{\underline{$0,72284\pm0,16996 $ }} \\
\hline
\tiny{clipTest500} & \tiny{ $0,82347\pm0,13901 $ }&\tiny{ $0,84531\pm0,15070 $} &\tiny{ $ 0,75748\pm0,15382 $ } & \tiny{\underline{$0,85621\pm0,16137 $ }}\\
\hline 
\tiny{ClipTest704} &\tiny{ $0,80168\pm0,08349 $ }&\tiny{ \underline{$0,85520\pm0,06826 $}} &\tiny{ $ 0,57703\pm0,07959 $ } & \tiny{ $0,78256\pm0,09523 $  } \\
\hline
\tiny{ClipTest752}  &\tiny{\underline{ $ 0,73288\pm0,17742$ }}&\tiny{ $0,54861\pm0,15555 $} &\tiny{ $0,71413\pm0,13138  $ } & \tiny{  $0,59356\pm0,17804  $  } \\
\hline
\tiny{ClipTest803} &\tiny{ $0,86825\pm0,106833 $ }&\tiny{$0,87556\pm0,06896 $} &\tiny{ $ 0,73847\pm0,14879 $ } & \tiny{ \underline{$0,88170\pm0,10827 $  }} \\
\hline
\tiny{ClipTest849} &\tiny{ $0,75279\pm0,15518 $ }&\tiny{ \underline{ $0,91888\pm0,07070 $}} &\tiny{ $0,55145\pm0,12245 $ } & \tiny{  $0,91089\pm0,05217 $}  \\
\hline

\end{tabular}   
\end{table}
Proposed DeepRGB8K saliency model turns to be winner more systématically (6/12 clips) than each reference model.
\subsection{Discussion}
Visual saliency prediction with deep CNN is still a recent while intensive research. The major bottle-neck in it is the computation power and memory requirements. We have shown, that
a very large amount of iterations - hundreds of thousands are not needed for prediction of interesting patches in video frames. Indeed, to get better  maximal accuracy smaller amount 
of iterations is needed, and the maximal number of iterations can be limited (17400 in our case) accompanied by another data selection strategy: all data from training set are passed
before each validation iteration of the learning, see tables \ref{tab_accuracyResultFirstEXper}, \ref{tab_accuracyResultSecondEXper}. 
Next, we have shown that in case of a sufficient training set, adding primary motion features improves prediction accuracy up to $2\%$ in average on a very large data set (HOLLYWOOD
test) containing $257733$ video frames. 
Hence the deep CNN captures the sensitivity of Human Visual System to motion.

When applying a supervised learning approach to visual saliency prediction in video, one has to keep in mind that gaze-fixation maps, which serve for selection of training "salient" 
regions in video frames, not only express the "bottom-up" attention. 
Humans are attracted by stimuli, but in case of video when  understanding a visual scene
with time, they focus on the objects of interest, thus reinforcing the "top-down" mechanisms of visual attention\cite{New52}. Hence, the prediction of patches of interest by a 
supervised learning, we mix all mechanisms: bottom-up and top-down. 

In order to re-inforce the bottom-up sensitivity of HVS to contrasts, we completed the input data layers by specific
contrast features well studied in classical saliency prediction models. As we could not state the improvement of performance in prediction of saliency of patches in video frames in  average 
(see table \ref{tab_accuracyResultSecondEXper}) a more detailed experience clip - by- clip was performed on a sample of clips from HOLLYWOOD dataset when comparing resulting predicted saliency
maps. This series of experiments resumed in table \ref{comparaison_withReference}, shows that indeed adding features, expressing local color contrast slightly improves performances 
with regard to the reference
bottom-up spatial (GBVS, SignatureSal) and spatio-temporal models (Seo)). 
Hence, the mean improvement of the complete model with motion, contrast features and primary HSV colour pixel 
values with regard to Itti, Harell and Seo models are $0.00677$, $0.01560$, $0.15862$ respectively. 
\begin{table}[H]
\caption{The mean improvement  of the complete model.} 
\label{comparaison_withReference}
\begin{tabular}{ | c | c | c | c | }
\hline
\tiny{ $nbr\_frame$ }& \tiny{$\overline{\delta}$(DeepRGB8k - GBVS) } & \tiny{ $\overline{\delta}$(DeepRGB8k - SignatureSal)} & \tiny{ $\overline{\delta}$(DeepRGB8k - Seo) } \\ 
\hline
$ 1614 $ & $0,00677\pm0,16922$ &$0,01560\pm0,19025$ & $0,15862\pm0,21036$ \\
 \hline
\end{tabular}
 
\end{table}

\section{Conclusion}
Hence, in this paper, we proposed a deep convolutional network to predict salient areas (patches) in video content and built dense predicted visual saliency maps upon them. We built an 
adequate architecture on the basis of Caffe CNN. While the aspiration of the community consisted in the use of primary features such as RGB planes only for visual attention prediction  
in images, we have shown that for video, adding of features expressing sensitivity of the human visual system to residual motion, is important. 
Furthermore, we also completed the RGB pixel values by low-level features of contrast and colour which are easy to compute and have proven efficient in former spatio-temporal 
predictors of visual attention. 
The results are better, nevertheless, the gain is not strong. Therefore, it is clear that for further research it is important to better explore the link between known physiological
mechanisms of human vision and the design of a CNN. The central bias hypothesis namely needs to be explored. 

\section*{References}
\bibliographystyle{harvard}
\bibliography{mybibfile}

\end{document}